\documentclass[shortAfour,sageh,times]{sagej}
\pdfoutput=1
\usepackage{moreverb,url}

\def\volumeyear{2019}

\usepackage{graphicx}
\usepackage{algorithm}
\usepackage{algpseudocode}
\usepackage[verbose]{wrapfig}
\usepackage[caption=false]{subfig}
\usepackage{todonotes}

\usepackage{tabto}

\newcommand\CONDITION[2]%
  {\begin{tabular}[t]{@{}l@{}l@{}}
     #1&#2
  \end{tabular}%
  }
\algdef{SE}[WHILE]{While}{EndWhile}[1]%
  {\algorithmicwhile\ \CONDITION{#1}{\ \algorithmicdo}}%
  {\algorithmicend\ \algorithmicwhile}
\algdef{SE}[FOR]{For}{EndFor}[1]%
  {\algorithmicfor\ \CONDITION{#1}{\ \algorithmicdo}}%
  {\algorithmicend\ \algorithmicfor}
\algdef{S}[FOR]{ForAll}[1]%
  {\algorithmicforall\ \CONDITION{#1}{\ \algorithmicdo}}
\algdef{SE}[REPEAT]{Repeat}{Until}{\algorithmicrepeat}[1]%
  {\algorithmicuntil\ \CONDITION{#1}{}}
\algdef{SE}[IF]{If}{EndIf}[1]%
  {\algorithmicif\ \CONDITION{#1}{\ \algorithmicthen}}%
  {\algorithmicend\ \algorithmicif}%
\algdef{C}[IF]{IF}{ElsIf}[1]%
  {\algorithmicelse\ \algorithmicif\ \CONDITION{#1}{\ \algorithmicthen}}

\usepackage{amsmath}
\usepackage{amssymb}
\usepackage{amsfonts}

\usepackage{amsthm}
\newtheorem{theorem}{Theorem}
\newtheorem{definition}[theorem]{Definition}
\newtheorem{assumption}[theorem]{Assumption}
\newtheorem{lemma}[theorem]{Lemma}

\usepackage{xifthen}

\usepackage{xcolor}
\newcommand{\rev}[1]{#1}

\newcommand{\dale}{Dale~M\textsuperscript{c}Conachie}
\newcommand{\dmitry}{Dmitry~Berenson}

\DeclareMathOperator*{\argmin}{argmin}

\DeclareMathOperator*{\proj}{Proj}

\usepackage{mathtools}

\newcommand{\reals}{\mathbb{R}}

\newcommand{\se}[1]{SE(#1)}
\newcommand{\tanse}[1]{\mathfrak{se}(#1)}
\newcommand{\eye}{\mathbf{I}}

\newcommand{\pr}{\mathbb{P}}

\newcommand{\config}{q}
\newcommand{\Config}{Q}

\newcommand{\robotconfig}[1][]{\config_
    {\ifthenelse{\isempty{#1}}{r}{r,#1}}
}
\newcommand{\robotvelocity}{\dot \config_r}

\newcommand{\robotcommandvel}[1][]{\dot \config^{cmd}_
    {\ifthenelse{\isempty{#1}}{r}{r,#1}}
}
\newcommand{\robotactualvel}[1][]{\dot \config^{act}_
    {\ifthenelse{\isempty{#1}}{r}{r,#1}}
}
\newcommand{\robotcommandsequence}{\dot \Config^{cmd}_r}

\newcommand{\maxrobotvel}[1][]{\robotvelocity^%
    {\ifthenelse{\isempty{#1}}{\textrm{max}}{\textrm{max},#1}}
}

\newcommand{\eeposition}[1][]{\config_%
    {\ifthenelse{\isempty{#1}}{\textrm{xyz}}{\textrm{xyz},#1}}
}
\newcommand{\eepositiongoal}[1][]{
    {\ifthenelse{\isempty{#1}}{\eeposition^\textrm{goal}}{\eeposition[#1]^\textrm{goal}}}
}

\newcommand{\gripperindex}{g}

\newcommand{\gripperconfigspace}{\se3 \times \se3}

\newcommand{\maxgrippervel}[1][]{\dot \config_%
    {\ifthenelse{\isempty{#1}}{r,\textrm{max}}{r,\textrm{max},#1}}
}
\newcommand{\maxgrippervelservo}{\maxgrippervel[e]}
\newcommand{\maxgrippervelobstacle}{\maxgrippervel[o]}

\newcommand{\relaxeddistancematrix}{D}
\newcommand{\deformconfig}{\mathcal{P}}
\newcommand{\deformvelocity}{\dot \deformconfig}
\newcommand{\numdeformpoints}{P}
\newcommand{\deformconfigspacesize}{{3\numdeformpoints}}
\newcommand{\deformconfigspace}{\reals^\deformconfigspacesize}

\newcommand{\deformtarget}{\mathcal{T}}
\newcommand{\correspondences}{\mathcal{T}_c}
\newcommand{\numtargetpoints}{T}
\newcommand{\errorfunction}{\rho}
\newcommand{\terminationcondition}{\Omega}
\newcommand{\maxstretchfactor}{\lambda_s}
\newcommand{\stretchingcorrectionweightfactor}{\lambda_w}

\newcommand{\band}{B}
\newcommand{\bandlength}{L}
\newcommand{\maxbandlength}{L_\textrm{max}}
\newcommand{\bandspace}{\mathbb{B}}
\newcommand{\bandspacevalid}{\bandspace^{valid}}
\newcommand{\bandspaceinv}{\bandspace^{inv}}
\newcommand{\bandgoal}{\bandspace^\textrm{goal}}
\newcommand{\numbandpoints}{N_{b,t}}
\newcommand{\maxbandpoints}{N_b^\textrm{max}}

\newcommand{\controller}{C}

\newcommand{\deformablemodelbackwardfunction}{\psi}

\newcommand{\pseudoinverseweight}{W}
\newcommand{\obstacle}{\mathcal{O}}

\newcommand{\taskexecutiontime}{N_e}
\newcommand{\truemotion}{E}
\newcommand{\predictionhorizon}{N_p}
\newcommand{\bandlengthannealing}{\alpha}
\newcommand{\historywindow}{N_h}
\newcommand{\errorprogressthreshold}{\beta_e}
\newcommand{\motionprogressthreshold}{\beta_m}

\newcommand{\goalreachradius}{\delta_\textrm{goal}}
\newcommand{\rrtnode}{\config}
\newcommand{\rrtnodeset}{\mathcal{V}}
\newcommand{\rrtedgeset}{\mathcal{E}}

\newcommand{\cost}{\textrm{Cost}}
\newcommand{\nodesnew}{\rrtnodeset^\textrm{new}}
\newcommand{\edgesnew}{\rrtedgeset^\textrm{new}}
\newcommand{\goalbias}{\gamma_{gb}}

\newcommand{\qbias}{\config_r^\textrm{bias}}
\newcommand{\Qapproxgoal}{\mathcal{\Config}_r^\textrm{goal}}
\newcommand{\Qgoal}{\mathbb{\Config}^\textrm{goal}}

\newcommand{\bestneardist}{\delta_{BN}}

\newcommand{\Dnear}[1][]{D_%
    {\ifthenelse{\isempty{#1}}{\textrm{near}}{\textrm{near},#1}}
}
\newcommand{\Dmax}[1][]{D_%
    {\ifthenelse{\isempty{#1}}{\textrm{max}}{\textrm{max},#1}}
}
\newcommand{\qrand}{\config^\textrm{rand}}
\newcommand{\qnear}{\config^\textrm{near}}
\newcommand{\qnearapprox}{\tilde{\config}^\textrm{near}}
\newcommand{\Qnear}{\Config^\textrm{near}}
\newcommand{\qlast}{\config^\textrm{last}}

\newcommand{\qinit}{\config^\textrm{init}}
\newcommand{\banddistscale}{\lambda_b}

\newcommand{\innerballsize}{\delta_\theta}
\newcommand{\ballseparation}{\delta_c}
\newcommand{\hyperball}{\mathcal{B}}

\newcommand{\cspace}{\mathbb{C}}
\newcommand{\cfree}{\cspace^{valid}}
\newcommand{\cinv}{\cspace^{inv}}
\newcommand{\cspacepath}{\pi}
\newcommand{\rpath}{\cspacepath^{ref}}
\newcommand{\setofpaths}{\mathbb{T}}

\newcommand{\selectprobability}{\gamma_k^{(i)}}
\newcommand{\selectbound}{\gamma_k^{(\infty)}}
\newcommand{\propagationprobability}{\rho_k^{(i)}}

\usepackage{makecell}
\usepackage{pbox}

\usepackage{silence}
\usepackage{placeins}

\begin{document}

\runninghead{M\textsuperscript{c}Conachie et. al.}

\title{Manipulating Deformable Objects by Interleaving Prediction, Planning, and Control}

\author{\dale\affilnum{1}, Andrew~Dobson\affilnum{1}, Mengyao~Ruan\affilnum{1}, and \dmitry\affilnum{1}}

\affiliation{\affilnum{1}University of Michigan, USA}

\corrauth{\dale, 
Robotics Institute,
University of Michigan,
1301 Beal Avenue,
Ann Arbor, MI, USA,
48109.}
\email{dmcconac@umich.edu}

\begin{abstract}
We present a framework for deformable object manipulation that interleaves planning and control, enabling complex manipulation tasks without relying on high-fidelity modeling or simulation. The key question we address is when should we use planning and when should we use control to achieve the task? Planners are designed to find paths through complex configuration spaces, but for highly underactuated systems, such as deformable objects, achieving a specific configuration is very difficult even with high-fidelity models. Conversely, controllers can be designed to achieve specific configurations, but they can be trapped in undesirable local minima due to obstacles. Our approach consists of three components: (1) A global motion planner to generate gross motion of the deformable object; (2) A local controller for refinement of the configuration of the deformable object; and (3) A novel deadlock prediction algorithm to determine when to use planning versus control. By separating planning from control we are able to use different representations of the deformable object, reducing overall complexity and enabling efficient computation of motion. We provide a detailed proof of probabilistic completeness for our planner, which is valid despite the fact that our system is underactuated and we do not have a steering function. We then demonstrate that our framework is able to successfully perform several manipulation tasks with rope and cloth in simulation which cannot be performed using either our controller or planner alone. These experiments suggest that our planner can generate paths efficiently, taking under a second on average to find a feasible path in three out of four scenarios. We also show that our framework is effective on a 16 DoF physical robot, where reachability and dual-arm constraints make the planning more difficult.
\end{abstract}

\maketitle

\section{Introduction}

Examples of deformable object manipulation range from domestic tasks like folding clothes to time and safety critical tasks such as robotic surgery. One of the challenges in planning for deformable object manipulation is the high number of degrees of freedom involved; even approximating the configuration of a piece of cloth in 3D with a 4 $\times$ 4 grid results in a 48 degree of freedom configuration space. In addition, the dynamics of the deformable object are difficult to model \citep{Essahbi2012}; even with high-fidelity modeling and simulation, planning for an individual task can take hours \citep{Bai2016}. Local controllers on the other hand are able to very efficiently generate motion, however, they are only able to successfully complete a task when the initial configuration is in the ``attraction basin'' of the goal \citep{Berenson2013,McConachie2018}.

\begin{figure*}[t]
    \centering
    \includegraphics[trim={5cm 6cm 3cm 4cm},clip,width=0.24\textwidth,height=1.5in,keepaspectratio]{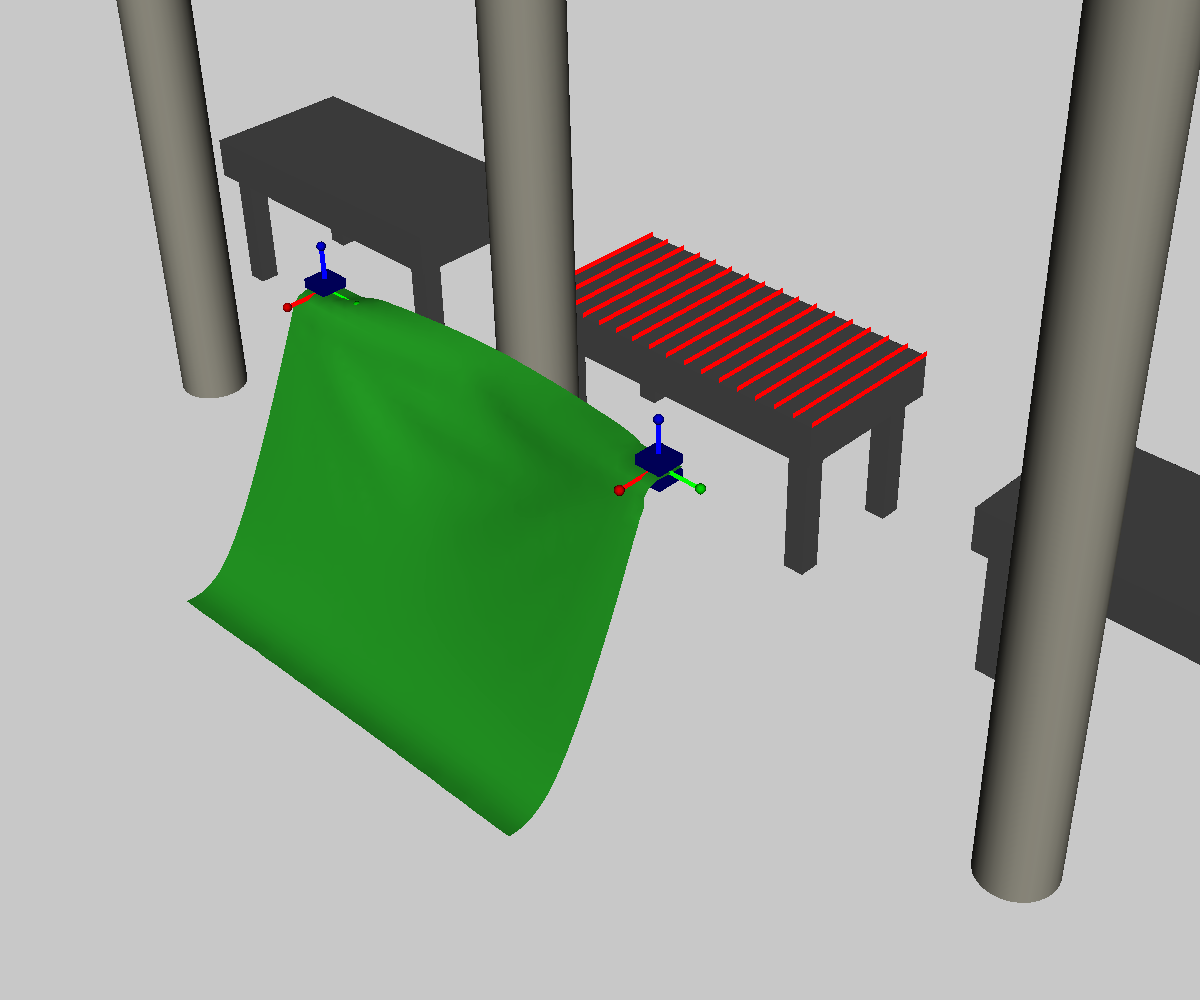}
    \hfill
    \includegraphics[trim={2cm 3cm 6cm 7cm},clip,width=0.24\textwidth,height=1.5in,keepaspectratio]{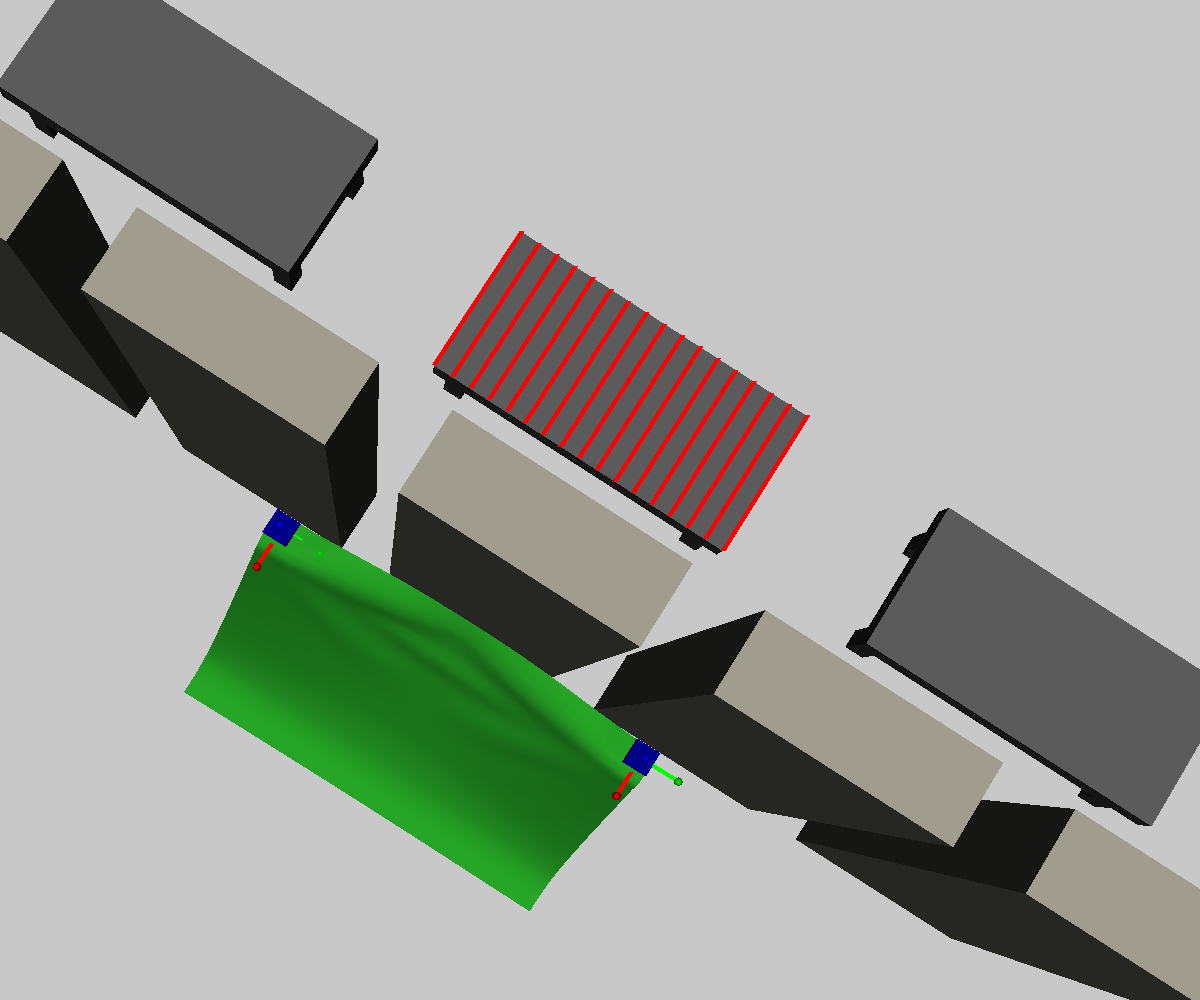}
    \hfill
    \includegraphics[trim={0cm 1cm 0cm 1cm},clip,width=0.24\textwidth,height=1.5in,keepaspectratio]{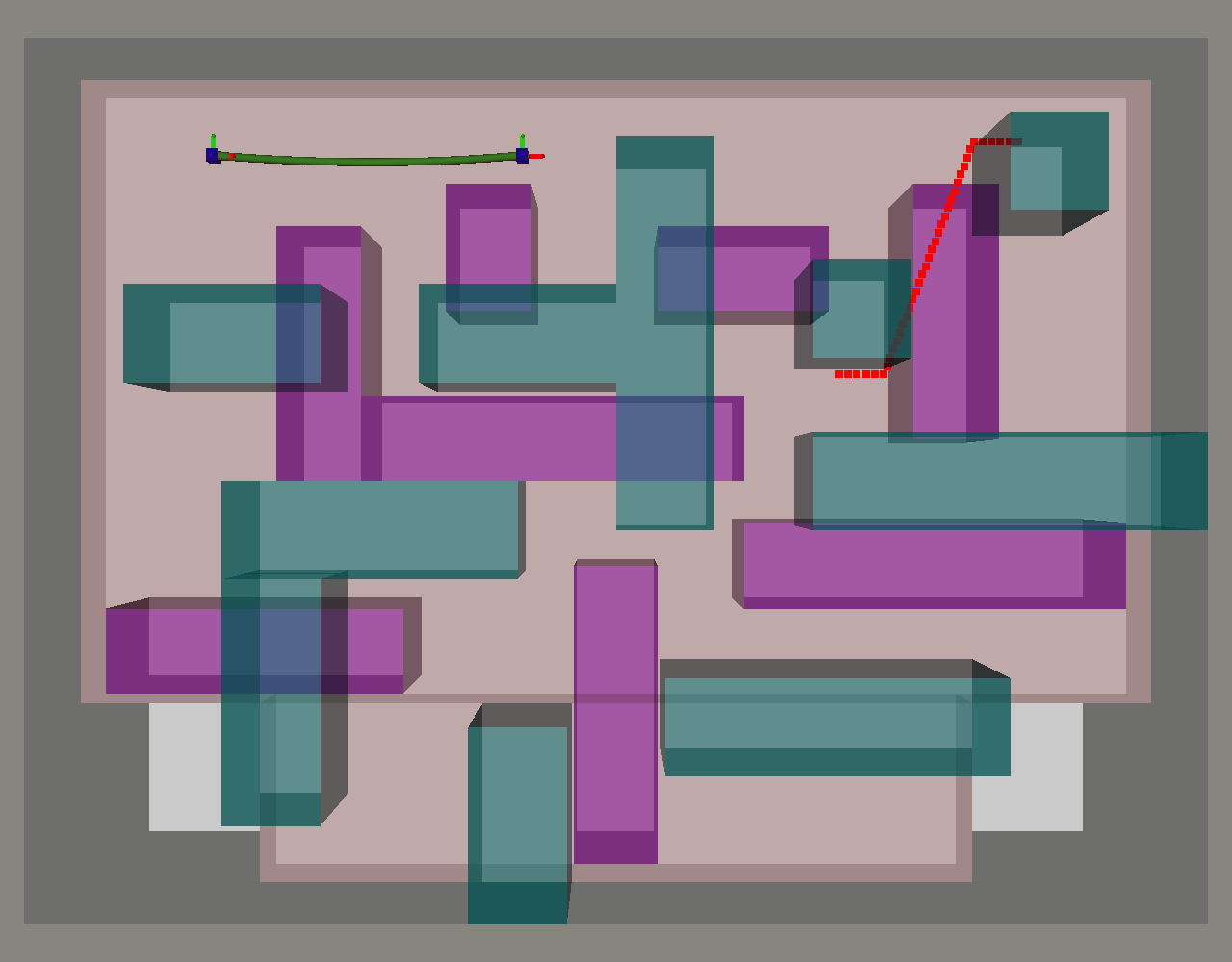}
    \hfill
    \includegraphics[trim={1.5cm 0cm 0.6cm 0cm},clip,width=0.24\textwidth,height=1.5in,keepaspectratio]{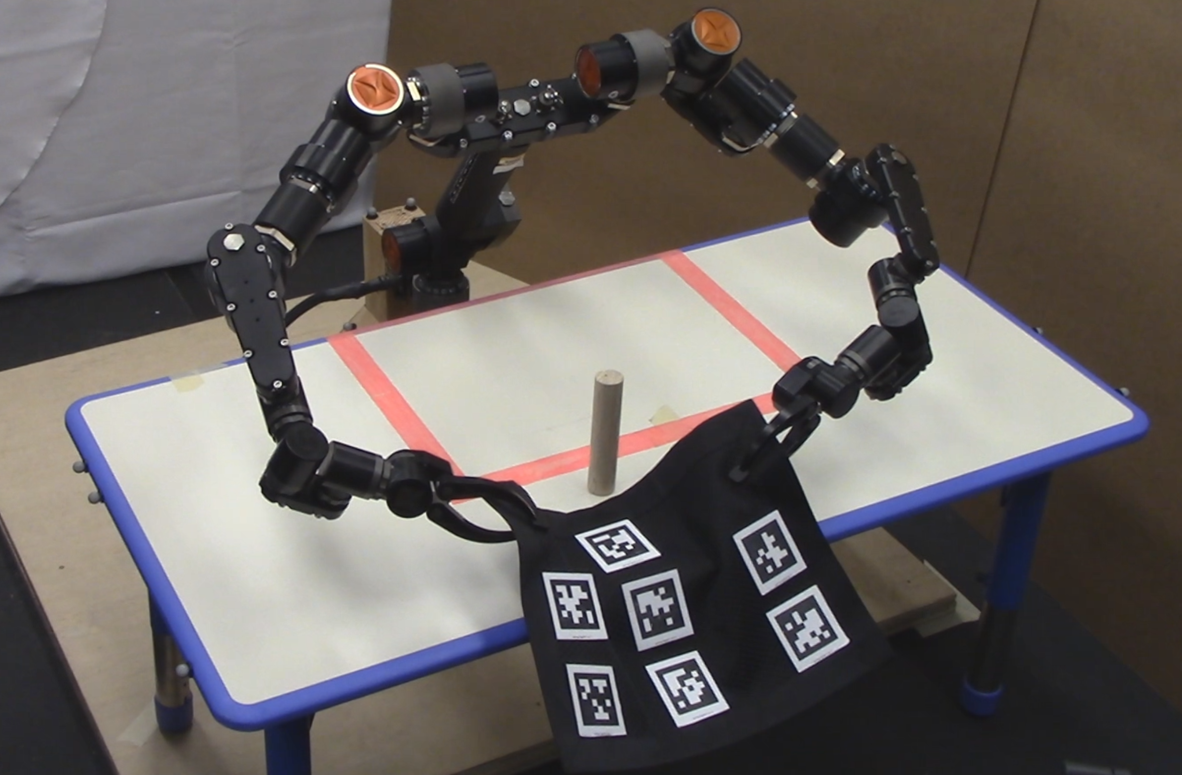}
    \caption{Four example manipulation tasks for our framework. In the first two tasks, the objective is to cover the surface of the table (indicated by the red lines) with the cloth (shown in green). In the first task, the grippers (shown in blue) can freely move however the cloth is obstructed by a pillar. In the second task, the grippers must pass through a narrow passage before the table can be covered. In the third task, the robot must navigate a rope (shown in green in the top left corner) through a three-dimensional maze before covering the red points in the top right corner. The maze consists of top and bottom layers (purple and green, respectively). The rope starts in the bottom layer and must move to the target on the top layer through an opening (bottom left or bottom right). For the fourth task, the physical robot must move the cloth from the far side of an obstacle to the region marked in pink near the base of the robot.}
    \label{fig:example_tasks}
\end{figure*}

The central question we address in this work is how can we combine the strengths of global planning with the strengths of local control while mitigating the weakness of each? We propose a framework for interleaving planning and control which uses global planning to generate gross motion of the deformable object, and a local controller to refine the configuration of the deformable object within the local neighborhood. By separating planning from control we are able to use different representations of the deformable object, each suited to efficient computation for their respective roles. In order to determine when to use each component, we introduce a novel deadlock prediction algorithm that is inspired by topologically-based motion planning methods \citep{Bhattacharya2012,Jaillet2008}. By answering the question ``Will the local controller get stuck?'' we can predict if the local controller will be unable to achieve the task from the current configuration. If we predict that the controller will get stuck we can then invoke the global planner, moving the deformable object into a new neighbourhood from which the local controller may be able to succeed. The key to our efficient prediction is forward-propagating only the stretching constraint, assuming the object will otherwise comply to contact.

We seek to solve problems for one-dimensional and two-dimensional deformable objects (i.e. rope and cloth) where we need to arrange the object in a particular way (e.g. covering a table with a tablecloth) but where there is also complex environment geometry preventing us from directly completing the task. While we cannot claim to solve all problems in this class (in particular in environments where the deformable object can be snagged), we can still solve practical problems where the path of the deformable object is obstructed by obstacles. In this work we restrict our focus to controllers of the form described in Sec.~\ref{sec:local_control}, and tasks suited to these controllers. Examples of these types of tasks are shown in Fig.~\ref{fig:example_tasks}. In our experiments we show that this iterative method of interleaving planning and control is able to successfully perform several interesting tasks where our planner or controller alone are unable to succeed.

Our contributions are: (1) A novel deadlock prediction algorithm to determine when a global planner is needed; (2) An efficient and probabistically-complete global planner for rope and cloth manipulation tasks; and (3) A framework to combine local control and global motion planning to leverage the strengths of each while mitigating their weaknesses. We present experiments in both a simulated environment and on a physical robot (Fig.~\ref{fig:example_tasks}). Our results suggest that our planner can efficiently find paths, taking under a second on average to generate a feasible path in three out of four simulated scenarios. The physical experiment shows that our framework is able to effectively perform tasks in the real world, where reachability and dual-arm constraints make the planning more difficult.


A preliminary version of this work was presented in \cite{McConachie2017}. This paper extends this work by adding an additional experiment on a physical robotic system as well as a proof of the probabilistic completeness of our planning method. We have also improved planning times with an improved goal bias method. We also include additional related work and an expanded discussion.
\section{Related Work}

Robotic manipulation of deformable objects has been studied in many contexts ranging from surgery to industrial manipulation (see \cite{Khalil2010} and \cite{Sanchez2018deformablesurvey} for extensive surveys). Below we discuss the most relevant methods to the work presented here, starting with methods of simulating and planning for deformable objects. We then discuss visual servoing and learning-based methods for similar tasks. In addition to previous work in deformable object manipulation, we also discuss related work in planning/control for robot arms and ways to consider topology in planning, which we draw from for our framework. We end with a discussion of probabilistic completeness and describe why previous methods to show this property do not apply, motivating our proof method.

Much work in deformable object manipulation relies on simulating an accurate model of the object being manipulated. Motivated by applications in computer graphics and surgical training, many methods have been developed for simulating string-like objects \citep{Bergou2008,Rungjiratananon2011} and cloth-like objects \citep{Baraff1998,Goldenthal2007}. The most common simulation methods use Mass-Spring models \citep{Gibson1997, Essahbi2012}, which are generally not accurate for large deformations \citep{Maris2010}, and Finite-Element (FEM) models \citep{Muller2002,Irving2004,Kaufmann2008}. FEM-based methods are widely used and physically well-founded, but they can be unstable when subject to contact constraints, which are especially important in this work. They also require significant tuning and are very sensitive to the discretization of the object. Furthermore, such models require knowledge of the physical properties of the object, such as it's Young's modulus and friction parameters, which we do not assume are known.

Motion planning for manipulation of deformable objects is an active area of research \citep{Jimenez2012}. \citet{Saha2008} present a Probabilistic Roadmap (PRM) \citep{Kavraki1996} that plans for knot-tying tasks with rope. \citet{Rodriguez2006} study motion planning in fully deformable simulation environments. Their method, based on Rapidly-exploring Random Trees (RRTs) \citep{LaValle2006}, applies forces directly to an object to move it through narrow spaces while using the simulator to compute the resulting deformations. \citet{Frank2011} presented a method that pre-computes deformation simulations in a given environment to enable fast multi-query planning. Other sampling-based approaches have also been proposed \citep{Anshelevich2000a,BurchanBayazit2002,Gayle2005,Lamiraux2001,Moll2006,Roussel2015}. However, all the above methods either disallow contact with the environment or rely on potentially time-consuming physical simulation of the deformable object, which is often very sensitive to physical and computational parameters that may be difficult to determine. In contrast our method uses simplified models for control and motion planning with far lower computational cost. In addition, the use of a local controller is not considered in the above methods, instead relying on a global planner (and thus implicitly the accuracy of the simulator) to generate a path that completes the entire task.

Model-based visual servoing approaches bypass planning entirely, and instead use a local controller to determine how to move the robot end-effector for a given task \citep{Hirai2000,Smolen2009,Wada2001}. Our recent work \citep{Berenson2013,McConachie2018} as well as \cite{Navarro-Alarcon2014,NavarroAlarcon2016,NavarroAlarcon2018} bypass the need for an explicit deformable object model, instead using approximations of the Jacobian to drive the deformable object to the attractor of the starting state. More recent work by \citet{Hu2018deformable_gpr} has enabled the use of Gaussian process regression while controlling a deformable object. Rather than using only a planner or only a controller, our framework uses both components, each when appropriate.

Approaches based on learning from demonstration avoid planning and deformable object modelling challenges entirely by using offline demonstrations to teach the robot specific manipulation tasks \citep{Huang2015,Schulman2016}; however, when a new task is attempted a new training set needs to be generated. In our application we are interested in a way to manipulate a deformable object without a high-fidelity model or training set available \textit{a priori}. For instance, imagine a robot encountering a new piece of clothing for a new task. While it may have models for previously-seen clothes or training sets for previous tasks, there is no guarantee that those models or training sets are appropriate for the new task.

\citet{Park2014Interleaving} considered interleaving planning and control for arm reaching tasks in rigid unknown environments. In their method, they assume an initially unknown environment in which they plan a path to a specific end-effector position. This path is then followed by a local controller until the task is complete, or the local controller gets stuck. If the local controller gets stuck, then a new path is planned and the cycle repeats. In contrast, our controller is performing the task directly rather than following a planned reference trajectory, incorporating deadlock prediction into the execution loop, while our global planner is planning for both the robot motion as well as the deformable object stretching constraint.

Our planning method has some similarity to topological \citep{Bhattacharya2012,Jaillet2008} and tethered robot \citep{Brass2015,SoonkyumKim2015} planning techniques; these methods use the topological structure of the space to define homotopy classes, either as a direct planning goal, or as a way to help inform planning in the case of tethered robots. Planning for some deformable objects, in particular rope or string, can be viewed as an extension of the tethered robot case where the base of the tether can move. This extension, however, requires a very different approach to homotopy than is commonly used, particularly when working in three-dimensional space instead of a planar environment. In our work we use \textit{visiblity deformations} from \cite{Jaillet2008} as a way to encode homotopy-like classes of configurations.

Previous approaches to proving probabilistic completeness for efficient planning of underactuated systems rely on the existence of a steering function to move the system from one region of the state space to another, or choosing controls at random \citep{LaValle2001,Karaman2013,Kunz2015,LiAOKP2016}. For deformable objects, a computationally-efficient steering function is not available, and using random controls can lead to prohibitively long planning times. \citet{Roussel2015} bypass this challenge by analyzing completeness in the submanifold of quasi-static contact-free configurations of a extensible elastic rods. In contrast, we show that our method is probabilistically complete even when contact between the deformable object and obstacles is considered along the path. Note that it is especially important to allow contact at the goal configuration of the object to achieve coverage tasks. \citet{LiAOKP2016} present an efficient asymptotically-optimal planner which does not need a steering function, however, they do rely on the existence of a contact free trajectory where every point in the trajectory is in the interior of the valid configuration space. Our proof of probabilistic completeness is based on \citet{LiAOKP2016}, but we allow for the deformable object to be in contact with obstacles along a given trajectory.

\section{Problem Statement}
\label{sec:main_problem_statement}


Define the robot configuration space to be $\cspace_r$. We assume that the robot configuration can be measured exactly. Denote an individual robot configuration as $\robotconfig \in \cspace_r$. This set can be partitioned into a valid and invalid set. The valid set is referred to as $\cfree_r$, and is the set of configurations where the robot is not in collision with the static geometry of the world. The invalid set is referred to as $\cinv_r = \cspace_r \setminus \cfree_r$.

We assume that our model of the robot is purely kinematic, with no higher order dynamics. We assume that the robot has two end-effectors that are rigidly attached to the object. The configuration of a deformable object is a set $\deformconfig \subset \reals^3$ of $\numdeformpoints = | \deformconfig |$ points. We assume that we have a method of sensing $\deformconfig$. The rest of the environment is denoted $\obstacle$ and is assumed to be both static, and known exactly. We assume that the robot moves slowly enough that we can treat the combined robot and deformable object as quasi-static. Let the function $f(\robotconfig, \deformconfig, \robotvelocity)$ map the system configuration $(\robotconfig, \deformconfig)$ and robot movement $\robotvelocity$ to the corresponding deformable object movement $\deformvelocity$. \rev{We assume that the deformable object will be damaged if it is stretched beyond a factor $\maxstretchfactor$ from the relaxed state. Let $\relaxeddistancematrix \in \reals^{\numdeformpoints^2}$ be the symmetric matrix of pairwise distances between all points of $\deformconfig$ in its relaxed state. We assume that there are no other deformable object properties (such as bending energy) that are relevant to the task.}

We define a task based on a set of $\numtargetpoints$ target points $\deformtarget \subset \reals^3$, a function $\errorfunction: \deformconfig \times \deformtarget \rightarrow \reals^{\geq 0}$, which measures the alignment error between $\deformconfig$ and $\deformtarget$, and a termination function $\terminationcondition (\deformconfig)$ which indicates if the task is finished. Let a robot controller be a function $\controller \left(\robotconfig, \deformconfig, \deformtarget \right)$\footnote{A specific controller may have additional parameters (such as gains in a PID controller), but we do not include such parameters here to keep $\controller(\dots)$ in a more general form.} which maps the system state $\left( \robotconfig, \deformconfig \right)$ and alignment targets $\deformtarget$ to a desired robot motion $\robotcommandvel$. In this work we restrict our discussion to tasks and controllers of the form introduced in our previous work (\cite{Berenson2013,McConachie2018}); these controllers are local, i.e. at each time $t$ they choose an incremental movement $\robotcommandvel$ which reduces the alignment error as much as possible at time $t + 1$. 


The problem we address in this work is how to find a sequence of $\taskexecutiontime$ robot commands $\{ \robotcommandvel[1], \dots, \robotcommandvel[\taskexecutiontime] \} = \robotcommandsequence$ such that each motion is feasible, i.e. it should not bring the grippers into collision with obstacles, should not cause the object to stretch excessively, and should not exceed the robot's maximum velocity $\maxrobotvel$. Let these feasibility constraints be represented by $A(\robotvelocity) = 0$. Then the problem we seek to solve is:
\begin{equation}
    \begin{aligned}
        & \text{find}   & & \robotcommandsequence \\
        & \text{s.t.}   & & \terminationcondition(\deformconfig_{\taskexecutiontime}) = \texttt{true} \\
        &               & & A(\robotcommandvel[t])                                    = 0, \; t = 1, \dots, \taskexecutiontime
    \end{aligned}
    \label{eqn:main_problem_statement}
\end{equation}
where $\deformconfig_{\taskexecutiontime}$ is the configuration of the deformable object after executing $\robotcommandsequence$.

Solving this problem directly is impractical in the general case for two major reasons. First, modeling a deformable object accurately is very difficult in the general case, especially if it contacts other objects or itself. Second, even given a perfect model, computing precise motion of the deformable object requires physical simulation, which can be very time consuming inside a planner/controller where many potential movements need to be evaluated. We seek a method which does not rely on high-fidelity modelling and simulation; instead we present a framework combining both global planning and local control to leverage the strengths of each in order to efficiently perform the task.

\section{Interleaving Planning and Control}

\begin{figure}[t]
    \centering
    \includegraphics[width=\columnwidth]{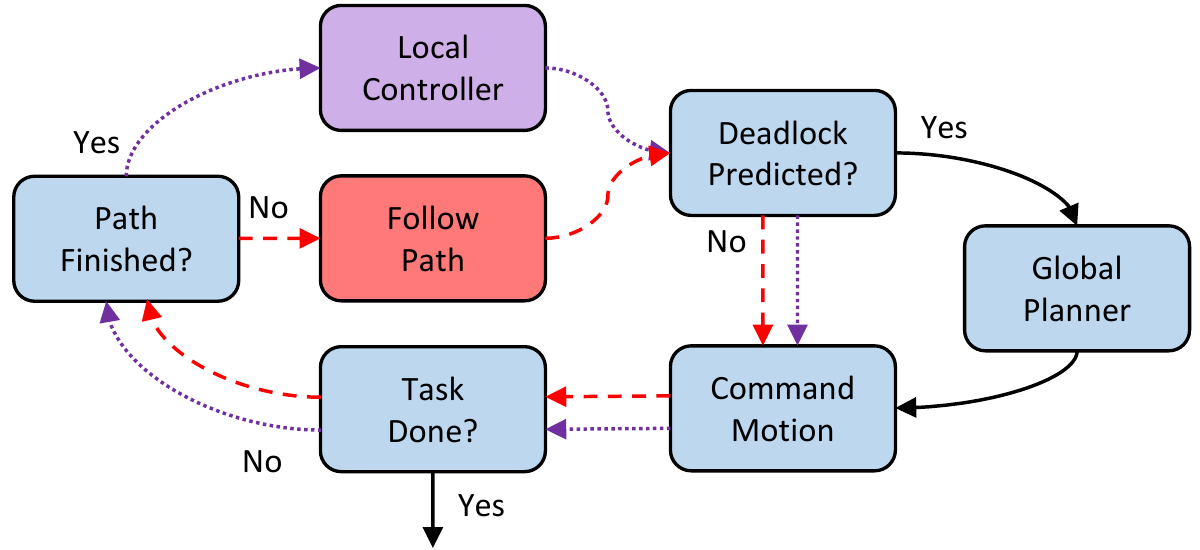}
    \caption{Block diagram showing the major components of our framework. On each cycle we use either the local controller (dotted purple arrows) or a planned path (dashed red arrows) to predict if the system will be deadlocked in the future, planning a new path is needed to avoid deadlock.}
    \label{fig:main_loop_diagram}
\end{figure}

Global planners are effective at finding paths through complex configuration spaces, but for highly underactuated systems such as deformable objects achieving a specific configuration is very difficult even with high-fidelity models; this means that we cannot rely on them to complete a task independent of a local controller. In order for the local controller to complete the task, the system must be in the correct basin of attraction. From this point of view it is not the planner's responsibility to complete a task but rather to move the system into the right basin for the local controller to finish the task. By explicitly separating planning from control we can use different representations of the deformable object for each component; this allows us to use a highly-simplified model of the deformable object for global planning to generate gross motion of the deformable object, while using an independent local approximation for the controller. The key question then is when should we use global planning versus local control?

Our framework can be broken down into three major components: (1) A global motion planner to generate gross motion of the deformable object; (2) A local controller for refinement of the configuration of the deformable object; and (3) A novel deadlock prediction algorithm to determine when to use planning versus control. Fig.~\ref{fig:main_loop_diagram} shows how these components are connected, switching between a local controller loop and planned path execution loop as needed. In the following sections we describe each component in turn, starting with the local controller.

\subsection{Local Control}
\label{sec:local_control}

The role of the local controller is not to perform the whole task, but rather to refine the configuration of the deformable object locally. For our local controller we use a controller of the form introduced in \cite{Berenson2013} and \cite{McConachie2018}. These controllers locally minimize error $\errorfunction$ while avoiding robot collision and excessive stretching of the deformable object.

An outline of how these controllers function is shown in Alg.~\ref{alg:local_controller}; first, for every target point $\deformtarget_i \in \deformtarget$ we define a workspace navigation function pointing towards $\deformtarget_i$ using Dijkstra's algorithm. This gives us the shortest collision-free path between any point in the workspace and the target point, as well as the distance travelled along that path. These navigation functions are used to define the best direction to move the deformable object $\deformvelocity_e$ and the relative importance of each part of the motion $\pseudoinverseweight_e$ in order to locally reduce error as much as possible at each timestep (Lines 1 and 2).
These error reduction terms are then combined \rev{using relative importance weight $\stretchingcorrectionweightfactor$} with stretching avoidance terms $\deformvelocity_s, \pseudoinverseweight_s$ to define the desired manipulation direction and importance weights $\deformvelocity_d, \pseudoinverseweight_d$ at each timestep (Lines 3 and 4). \rev{If these terms conflict, then stretching correction takes precedence.}
We then find the best robot motion to achieve the desired deformable object motion, while preventing collision between the robot and obstacles (Line 5).

Given the current system state $(\robotconfig, \deformconfig)$ FindBestRobotMotion$(\robotconfig, \deformconfig, \deformvelocity_d, \pseudoinverseweight_d)$ is solving the following problem:
\begin{equation}
\begin{aligned}
    & \argmin_{\robotvelocity } 
        & & \| f(\robotconfig, \deformconfig, \robotvelocity) - \deformvelocity_d \|_{\pseudoinverseweight_d} \\
    &\text{subject to}
        & & \| \robotvelocity \| \leq \maxrobotvel \\
    &   & & \left(\robotconfig + \robotvelocity\right) \in \cfree_r \enspace .
\end{aligned}
\label{eqn:controller_minimization_problem}
\end{equation}
How Eq.~\eqref{eqn:controller_minimization_problem} is solved depends on the particular robot; details for each function in Alg.~\ref{alg:local_controller} are in Appendix~\ref{apx:local_control}.

\begin{algorithm}[t]
    \caption{LocalController$(\robotconfig, \deformconfig, \deformtarget, \relaxeddistancematrix, \maxstretchfactor, \stretchingcorrectionweightfactor)$}
    \begin{algorithmic}[1]
        \State $\correspondences \gets$ CalculateCorrespondences$(\deformconfig_t, \deformtarget)$
        \State $\deformvelocity_e, \pseudoinverseweight_e \gets$ FollowNavigationFunction$(\deformconfig_n, \correspondences)$
        \State $\deformvelocity_s, \pseudoinverseweight_s \gets$ StretchingCorrection$(\relaxeddistancematrix, \maxstretchfactor, \deformconfig)$
        \State $\deformvelocity_d, \pseudoinverseweight_d \gets$ CombineTerms$(\deformvelocity_e, \pseudoinverseweight_e, \deformvelocity_s, \pseudoinverseweight_s, \stretchingcorrectionweightfactor)$
        \State $\robotcommandvel \gets$ FindBestRobotMotion$(\robotconfig, \deformconfig, \deformvelocity_d, \pseudoinverseweight_d)$
    \end{algorithmic}
    \label{alg:local_controller}
\end{algorithm}

\begin{figure}[t]
    \centering
    \includegraphics[width=\columnwidth]{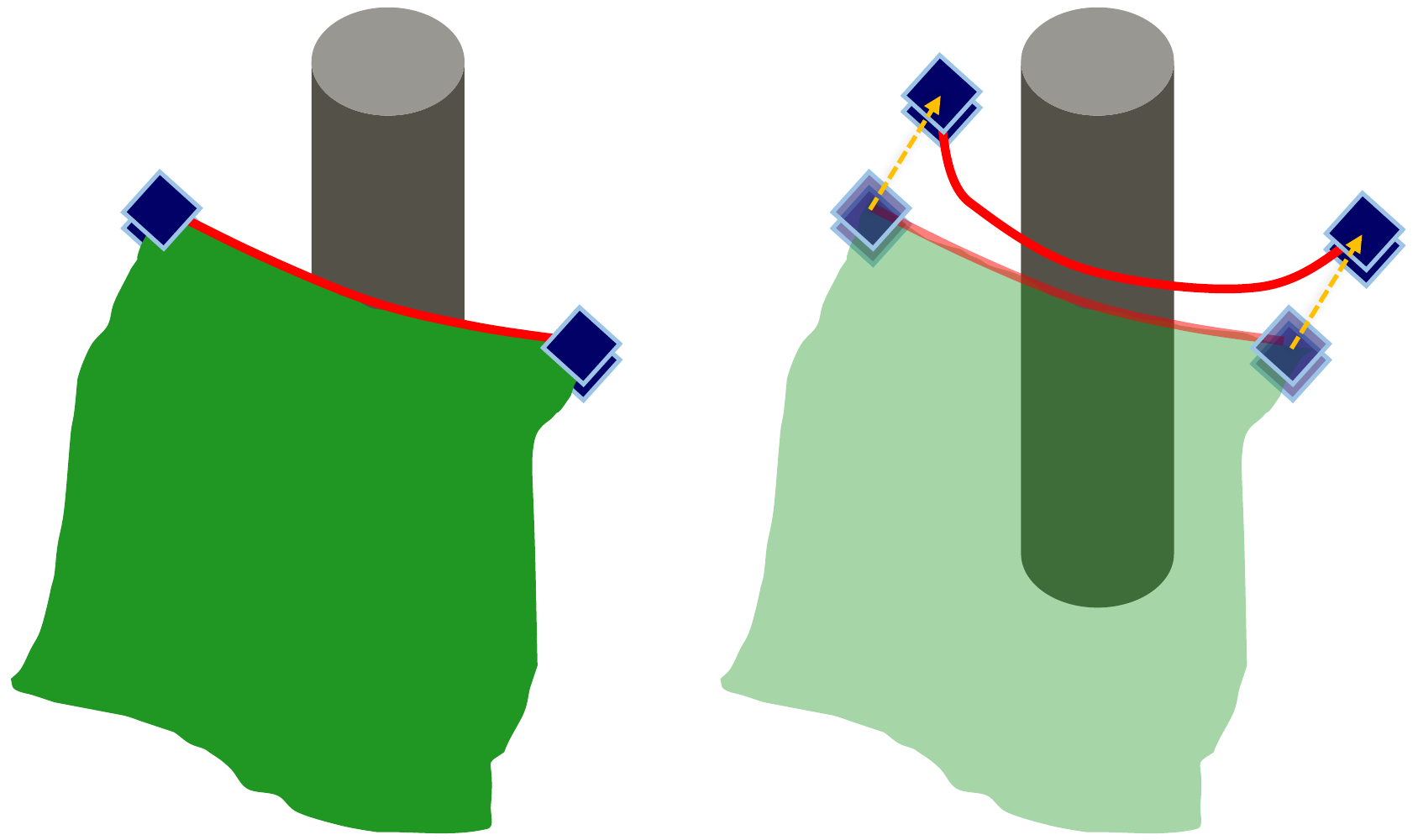}
    \caption{Motivating example for deadlock prediction. The local controller moves the grippers on opposite sides of an obstacle, while the geodesic between the grippers (red line) cannot move past the pole, eventually leading to overstretch or tearing of the deformable object if the robot does not stop moving towards the goal.}
    \label{fig:overstretch_example}
\end{figure}

An important limitation of this approach is that the individual navigation functions are defined and applied independently of each other; this means that the navigation functions that are combined to define the direction to move the deformable object can cause the controller to move the end effectors on opposite sides of an obstacle, leading to poor local minima, i.e. becoming stuck. Figure~\ref{fig:overstretch_example} shows our motivating example of this type of situation. Other examples of this kind of situation are shown in Section \ref{sec:simulation_experiments}. In addition, while this local controller prevents collision between the robot and obstacles, it does not explicitly have any ability to \textit{go around} obstacles.

In order to address these limitations we introduce a novel deadlock prediction algorithm to detect when the system $(\robotconfig[t], \deformconfig_t)$ is in a state that will lead to deadlock (i.e. becoming stuck) if we continue to use the local controller.

\subsection{Predicting Deadlock}
\label{sec:predicting_deadlock}

Predicting deadlock is important for two reasons; first we do not want to waste time executing motions that will not achieve the task. Second, we want to avoid the computational expense of planning our way out of a cul-de-sac after reaching a stuck state. By predicting deadlock before it happens we address both of these concerns. The key idea is to detect situations similar to Figure~\ref{fig:overstretch_example} where the local controller will wrap the deformable object around an obstacle without completing the task. We also need to detect situations where no progress can be made due to an obstacle directly in the path of the desired motion of the robot.

Let $\truemotion(\robotconfig, \deformconfig, \robotcommandvel) = \robotactualvel$ be the true motion of the robot when $\robotcommandvel$ is executed for unit time; in this section we will be predicting the future state of the system, thus it is not sufficient to consider only $\robotcommandvel$, we must also consider $\robotactualvel$.  Modelling inaccuracies as well as the deformable object being in contact can lead to meaningful differences between $\robotcommandvel$ and $\robotactualvel$. Specifically, when a deformable object is in contact with the environment, tracking $\robotcommandvel$ perfectly may lead to a constraint violation (i.e. overstretch or tearing of the deformable object).

We consider a controller to be deadlocked if the commanded motion produces (nearly) no actual motion, and the task termination condition is not met:
\begin{equation}
    \begin{split}
        \| \robotactualvel[t] \|               &\approx 0 \\
        \terminationcondition(\deformconfig_t) &= \texttt{false}.
        \label{eqn:stuck}
    \end{split}
\end{equation}
In general we cannot predict if the system will get stuck in the limit; to do so would require a very accurate simulation of the deformable object. Instead we predict if the system will get stuck within a prediction horizon $\predictionhorizon$ timesteps. We divide our deadlock prediction algorithm into three parts and discuss each in turn: 1) estimating gross motion; 2) predicting overstretch; and 3) progress detection.

\subsubsection{Estimating Gross Motion:}

\begin{figure}
    \centering
    \includegraphics[width=\columnwidth]{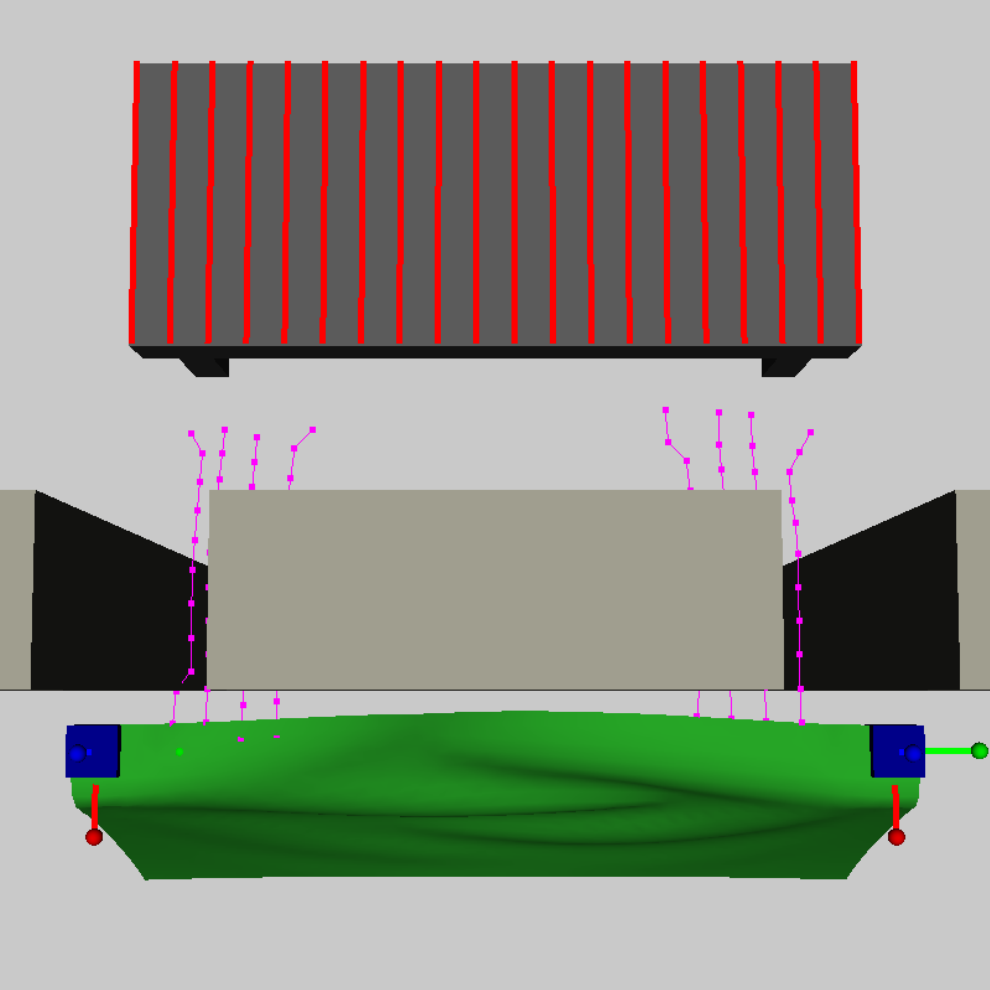}
    \caption{Example of estimating the gross motion of the deformable object for a prediction horizon $\predictionhorizon = 10$. The magenta lines start from the points of the deformable object that are closest to the target points (according to the navigation function). These lines show the paths those points would follow to reach the target when following the navigation function.}
    \label{fig:gross_deformable_motion}
\end{figure}

The idea central to our prediction (Alg.~\ref{alg:predict_deadlock}) is that while we may not be able to determine precisely how a given controller will steer the system, we can capture the gross motion of the system and estimate if the controller will be deadlocked. We split the prediction into two parts; first we assume that controller $\controller$ is able to manipulate the deformable object with a reasonable degree of accuracy within a local neighborhood of the current state. This allows us to approximate the motion of the deformable object by following the task-defined navigation functions for each $\deformconfig_i \in \deformconfig$. Examples of this approximation are shown in Figure~\ref{fig:gross_deformable_motion}.

Next we use a simplified version of LocalController() which omits the stretching avoidance terms (Alg.~\ref{alg:local_controller} lines 3 and 4) to predict the commands sent to the robot. These terms are omitted as they can be sensitive to the exact configuration of the deformable object, which is not considered in this approximation. If we are executing a path then we can use the planned path directly to predict overstretch.

\begin{algorithm}[t]
\caption{PredictDeadlock$($\parbox[t]{1.5in}{$\errorfunction, \robotconfig[t], \deformconfig_t, \band_t, \deformtarget,$\\$\maxbandlength, \predictionhorizon,\textrm{Path})$}}
\begin{algorithmic}[1]
    \State ConfigHistory $\gets [\textrm{ConfigHistory}, q_t]$
    \State ErrorHistory $\gets [\textrm{ErrorHistory}, \errorfunction(\deformconfig_t)]$
    \State BandPredictions $\gets []$
    \State $\correspondences \gets$ CalculateCorrespondences$(\deformconfig_t, \deformtarget)$
    \For{$n = t, \dots, t + \predictionhorizon - 1$}
        \If {Path $\neq \emptyset$}
            \State $\deformvelocity_e, \pseudoinverseweight_e \gets$ FollowNavigationFunction$(\deformconfig_n, \correspondences)$
            \State $\deformconfig_{n + 1} \gets \deformconfig_n + \deformvelocity_e$
            \State $\robotcommandvel[n] \gets$ FindBestRobotMotion$($\parbox[t]{0.8in}{$\robotconfig[n], \deformconfig_{n},$ \\ $\deformvelocity_e, \pseudoinverseweight_e)$}
            \State $\robotconfig[n + 1] \gets \robotconfig[n] + \robotcommandvel[n]$
        \Else
            \State $\robotconfig[n + 1] \gets \robotconfig[n] + $ FollowPath(Path)
        \EndIf
        \State $\band_{n + 1} \gets$ ForwardPropagateBand$(\band_n, \robotconfig[n + 1])$
        \State BandPredictions $\gets [\textrm{BandPredictions}, \band_{n + 1}]$
    \EndFor
    \If {PredictOverstretch$(\textrm{BandPredictions},\ \maxbandlength)$ \textbf{or} \\NoProgress$($ConfigHistory, ErrorHistory$)$}
        \State \Return \texttt{true}
    \Else
        \State \Return \texttt{false}
    \EndIf
\end{algorithmic}
\label{alg:predict_deadlock}
\end{algorithm}

\begin{algorithm}[t]
\caption{ForwardPropagateBand$(\band, \robotconfig)$}
\begin{algorithmic}[1]
    \State $(p_0, p_1) \gets$ ForwardKinematics$(\robotconfig)$
    \State $\band \gets$ $[p_0, \band, p_1]$
    \State $\band \gets$ InterpolateBandPoints$(\band)$
    \State $\band \gets$ RemoveExtraBandPoints$(\band)$
    \State $\band \gets$ PullTight$(\band)$
    \State \Return $\band$
\end{algorithmic}
\label{alg:band_propogation}
\end{algorithm}

\subsubsection{Predicting Overstretch:}
\label{sec:overstretch}

Next we introduce the notion of a \textit{virtual elastic band} between the robot's end-effectors. This elastic band represents the shortest path through the deformable object between the end-effectors. The band approximates the constraint imposed by the deformable object on the motion of the robot; if the end-effectors move too far apart, then the elastic band will be too long, and thus the deformable object is stretched beyond a task-specified maximum stretching factor $\maxstretchfactor$. Similarly, if the elastic band gets caught on an obstacle and becomes too long, then the deformable object is also overstretched. By considering only the geodesic between the end-effectors, we are assuming that the rest of deformable object will comply to the environment, and does not need to be considered when predicting overstretch. The elastic band representation allows us to use a fast prediction method, but does not account for the part of the material that is slack. We discuss this trade-off further in Section \ref{sec:discussion}. This virtual elastic band is based on Quinlan's path deformation algorithm~\cite{Quinlan1994} and is used both in deadlock prediction as well as global planning (Sec.~\ref{sec:planning_goal} and Sec.~\ref{sec:global_planning})

Denote the configuration of an elastic band at time $t$ as a sequence of $\numbandpoints$ points $\band_t \subset \reals^3$. The number of points used to represent an elastic band can change over time, but for any given environment and deformable object there is an upper limit $\maxbandpoints$ on the number of points used. Define $\textrm{Path}(\band)$ to be the straight line interpolation of all points in $\band$. Define the length of a band to be the length of this straight line interpolation. At each timestep the elastic band is initialized with the shortest path between the end effectors through the deformable object, and then ``pulled'' tight using the internal contraction force described in ~\cite{Quinlan1994} \textsection5, and a hard constraint for collision avoidance. The endpoints of the band track the predicted translation of the end effectors (Alg.~\ref{alg:band_propogation}). This band represents the constraint that must be satisfied for the object not to tear. By considering only this constraint on the object in prediction, we are implicitly relying on the object to comply to contact as it is moved by the robot. We discuss the limitations of this assumption in the discussion (Sec.~\ref{sec:discussion}).

\begin{figure}
    \centering
    \includegraphics[width=\columnwidth]{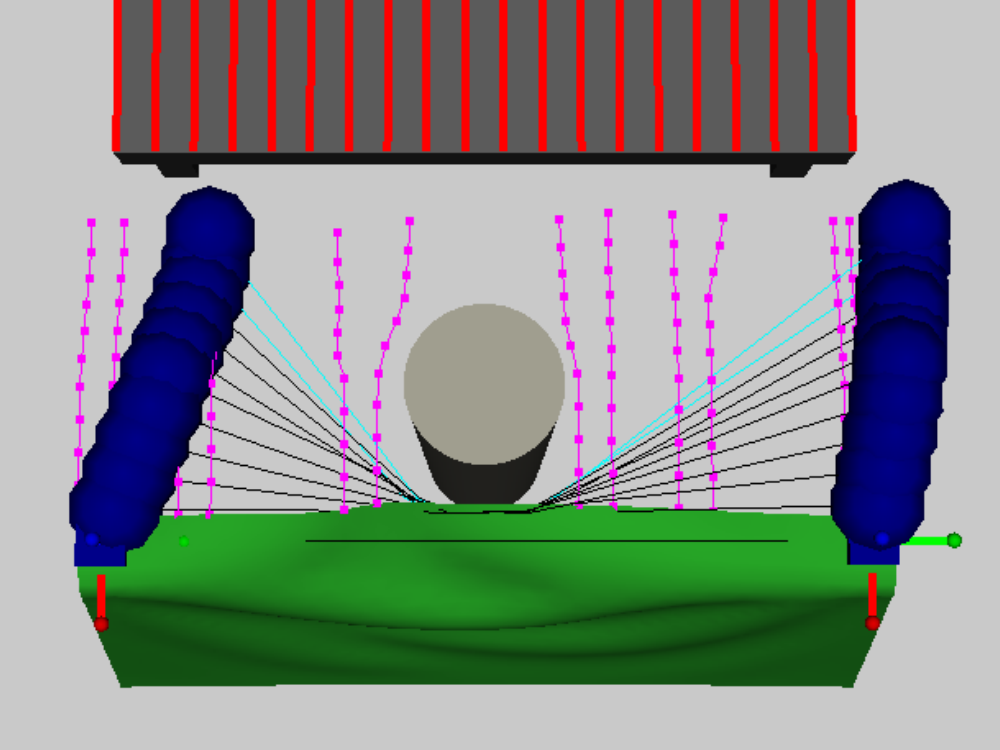}
    \caption{Estimated gross motion of the deformable object (magenta lines) and end effectors (blue spheres). The virtual elastic band (black lines) is forward propagated by tracking the end effector positions, changing to cyan lines when overstretch is predicted.}
    \label{fig:overstretch_predicted}
\end{figure}

Let $\bandlength_{t+n}$ be the length of the path defined by the virtual elastic band $\band_{t+n}$ at timestep $n$ in the future, and $\maxbandlength$ be the longest allowable band length. To use this length sequence to predict if the controller will overstretch the deformable object, we perform three filtering steps: an annealing low-pass filter, a filter to eliminate cases where the band is in freespace, and the detector itself which predicts overstretch. We use a low-pass annealing filter with annealing constant $\bandlengthannealing \in [0, 1)$ to mitigate the effect of numerical and approximation errors which could otherwise lead to unnecessary planning:
\begin{equation}
    \begin{split}
        \tilde \bandlength_{t + 1} &= \bandlength_{t + 1} \\
        \tilde \bandlength_{t + n} &= \bandlengthannealing \tilde \bandlength_{t + n - 1} + (1 - \bandlengthannealing) \bandlength_{t + n} \enspace ,  n = 2, \dots, \predictionhorizon \enspace .
    \end{split}
\end{equation}
Second, we discard from consideration any bands which are not in contact with an obstacle; we can eliminate these cases because our local controller includes an overstretch avoidance term which will prevent overstretch in this case in general. Last we compare the filtered length of any remaining band predictions to $\maxbandlength$; if after filtering, there is an estimated band length $\tilde \bandlength$ that is larger than $\maxbandlength$ then we predict that the local controller will be stuck. An example of this type of detection is shown in Figure~\ref{fig:overstretch_predicted}, where the local controller will wrap the cloth around the pole, eventually becoming deadlocked in the process.

\subsubsection{Progress Detection:}

Last, we track the progress of the robot and task error to estimate if the controller $\controller$ is making progress towards the task goal. This is designed to detect cases when the robot is trapped against an obstacle. Naively we could look for instances when $\robotactualvel = 0$ however due to sensor noise, actuation error, and using discrete math in a computer, we need to use a threshold instead. At the same time we want to avoid false positives, where the robot is moving slowly but task error is decreasing. To address these concerns we record the configuration of the robot (stored in ConfigHistory) and the task error (stored in ErrorrHistory) every time we check for deadlock, and introduce three parameters to control what it means to be making progress: history window $\historywindow$, error improvement threshold $\errorprogressthreshold$, and configuration distance threshold $\motionprogressthreshold$. If over the last $\historywindow$ timesteps, the improvement in error is less than $\errorprogressthreshold$, and the robot has moved less than $\motionprogressthreshold$, then we predict that the controller will not be able to reach the goal from the current state and trigger global planning.

\subsection{Setting the Global Planning Goal}
\label{sec:planning_goal}

In order to enable efficient planning, we need to approximate the configuration of the deformable object in a way that captures the gross motion of the deformable object without being prohibitively expensive to use. We use the same approach from Sec.~\ref{sec:overstretch}, but the interpretation in this use is slightly different; the virtual elastic band is a proxy for the leading edge of the deformable object. \rev{To define the leading edge, we again use the geodesic between the grippers.} In this way we can plan to move the deformable object to a different part of the workspace without needing to simulate the entire deformable object, instead the deformable object conforms to the environment naturally.

In order to make progress towards achieving the task, we want to set the goal for the global planner to be a configuration that we have not explored with the local controller. We do so in two parts; we find the set of all target points $\deformtarget_U$ which are contributing to task error, split these points into two clusters, and use the cluster centers to define the goal region of the end effectors, $\eepositiongoal$; any end-effector position within a task-specified distance $\goalreachradius$ is considered to have reached the end-effector goal (Alg.~\ref{alg:call_global_planner} lines 1-3). Second, we set the goal configuration of the virtual elastic band to be any configuration that is not similar to a \textit{blacklist} of virtual elastic bands. This blacklist is the set of all band configurations from which we predicted that the local controller would be deadlocked in the future (Sec.~\ref{sec:predicting_deadlock}).

To define similarity we use Jaillet and Sim\'{e}on's \textit{visibility deformation} definition to compare two virtual elastic bands (\cite{Jaillet2008}). Intuitively two virtual elastic bands are similar if you can sweep a straight line connecting the two bands from the start points to the end points of the two bands without intersecting an obstacle. Unlike the original use, we do not constrain the start and end points of each path to match, but the algorithm is identical. We use this as a heuristic to find states that are dissimilar from states where we have already predicted that the local controller would be deadlocked. Let VisCheck$(\band,\textrm{Blacklist}) \rightarrow \{0, 1\}$ denote this visibility deformation check, returning $1$ if $\band$ is similar to a band in the blacklist and $0$ otherwise. Then
\begin{equation}
    \bandgoal = \{\band \mid \text{VisCheck}(\band, \textrm{Blacklist}) = 0\}
    \label{eqn:bandgoal}
\end{equation}
is the set of all virtual elastic bands that are dissimilar to the Blacklist.

Combined, $\eepositiongoal, \goalreachradius$, and $\bandgoal$ define what it means for the planner to have found a path to the goal (Alg.~\ref{alg:goal_check}); the end-effectors must be in the right region, and the virtual elastic band must be dissimilar to any band in the Blacklist.

\begin{algorithm}[t]
\caption{PlanPath$($\parbox[t]{1.8in}{$\robotconfig[t], \deformconfig_t, \band_t, \deformtarget, \goalreachradius,$\\
                                      $\maxbandlength, \bestneardist, \textrm{Blacklist})$}}
\begin{algorithmic}[1]
    \State $\deformtarget_U \gets$ UncoveredTargetPoints$(\deformconfig_t, \deformtarget)$
    \State $\eepositiongoal \gets$ ClusterCenters$(\deformtarget_U)$
    \State $\eepositiongoal \gets$ ProjectOutOfCollision$(\eepositiongoal)$
    \State $\bandgoal \gets \{\band \mid \text{VisCheck}(\band, \text{Blacklist}) = 0\}$
    \State Path $\gets$ RRT-EB$(\robotconfig[t], \band_t, \eepositiongoal, \goalreachradius, \bandgoal, \maxbandlength, \bestneardist)$
    \If {Path $\neq$ Failure}
        \State \Return ShortcutSmooth(Path)
    \Else
        \State \Return Failure
    \EndIf
\end{algorithmic}
\label{alg:call_global_planner}
\end{algorithm}

\begin{algorithm}[t]
\caption{GoalCheck$(\rrtnodeset, \eepositiongoal, \goalreachradius, \bandgoal)$}
\begin{algorithmic}[1]
    \For {$\config_f = (\robotconfig, \band) \in \rrtnodeset$}
        \State $(p_0, p_1) \gets$ ForwardKinematics$(\robotconfig)$
        \If {$\|p_0 - \eepositiongoal[0] \| \leq \goalreachradius$ \textbf{and} \\
             $\|p_1 - \eepositiongoal[1] \| \leq \goalreachradius$ \textbf{and} $\band \in \bandgoal$}
            \State \Return 1
        \EndIf
    \EndFor
    \State \Return 0
\end{algorithmic}
\label{alg:goal_check}
\end{algorithm}

The combination of local control, deadlock prediction, and global planning are shown in the MainLoop function (Alg.~\ref{alg:mainloop}). Because the virtual elastic band is an approximation we need to predict deadlock while executing the planned path. We use the same prediction method for path execution as for the local controller. To set the maximum band length $\maxbandlength$ used by the global planner and the deadlock prediction algorithms, we calculate the geodesic distance between the grippers through the deformable object in its ``laid-flat'' state and scale it by the task specified maximum stretching factor $\maxstretchfactor$.

\section{Global Planning}
\label{sec:global_planning}

The purpose of the global planner is not to find a path to a configuration where the task is complete, but rather to move the system into a state from which the local controller can complete the task. Planning directly in configuration space of the full system $\cspace_r \times \deformconfigspace$ is not practical for two important reasons. First, this space is very high-dimensional and the system is highly underactuated. More importantly, to accurately know the state of the deformable object after a series of robot motions one would need a high-fidelity simulation that has been tuned to represent a particular task. We seek to plan paths very quickly without knowing the physical properties of a deformable object \textit{a priori}. The key idea that allows us to plan paths quickly is to consider only the constraint on robot motion that is imposed by the deformable object; i.e. the robot motion shall not tear or cause excessive stretching of the deformable object. We represent this constraint using a virtual elastic band and enforce the constraint that the band's length cannot exceed $\maxbandlength$.

\subsection{Planning Setup}


Denote the planning configuration space as $\cspace_f = \cspace_r \times \bandspace$. In order to split $\cspace_f$ into valid and invalid sets, we first define what it means for a band $\band \in \bandspace$ to be valid. A band $\band \in \bandspace$ is considered valid if the band is not overstretched and the path defined by $\band$ does not \textit{penetrate} an obstacle:
\begin{equation}
\begin{split}
    \bandspacevalid = \{ \band \mid \enspace &\textrm{Length}(\band) \leq \maxbandlength \textrm{ and } \\
                                             &\textrm{Path(\band)} \cap \textrm{Interior}(\obstacle) = \emptyset \} \enspace .
\end{split}
\end{equation}
Then the invalid set is $\bandspaceinv = \bandspace \setminus \bandspacevalid$. Similarly define $\cfree_f = \cfree_r \times \bandspacevalid$ and $\cinv_f = \cspace_f \setminus \cfree_f$.

\begin{algorithm}[t]
\caption{MainLoop$($\parbox[t]{2.1in}{$\deformtarget, \terminationcondition, \errorfunction, \deformconfig_\textrm{flat}, \maxstretchfactor, \stretchingcorrectionweightfactor,$ \\ $\predictionhorizon, \goalreachradius,\bestneardist)$}}
\begin{algorithmic}[1]
    \State $\relaxeddistancematrix \gets$ GeodesicDistanceBetweenEndEffectors$(\deformconfig_\textrm{flat})$
    \State $\maxbandlength \gets \maxstretchfactor \relaxeddistancematrix$
    \State Blacklist $\gets \emptyset$
    \State Path $\gets \emptyset$
    \State $t \gets 0$
    \State $\robotconfig[0] \gets$ SenseRobotConfig$()$
    \State $\deformconfig_0 \gets$ SensePoints$()$
    
    \While{$\neg \terminationcondition(\deformconfig_t)$}
        \State $\band_t \gets$ InitializeBand$(\deformconfig_t)$
        
        \If {PredictDeadlock$(\errorfunction, \robotconfig[t], \deformconfig_t, \band_t, \deformtarget,$\\\hphantom{PredictDeadlock$($}$\maxbandlength, \predictionhorizon,\textrm{Path})$}
            \State Blacklist $\gets$ Blacklist $\cup \{ \band_t \}$
            \State Path $\gets$ PlanPath$($\parbox[t]{2.3in}{$\robotconfig[t], \deformconfig_t, \band_t, \deformtarget, \goalreachradius,$\\
                                                             $\maxbandlength, \bestneardist, \textrm{Blacklist})$}
            \If {Path = Failure}
                \State \Return Failure
            \EndIf
        \EndIf
        
        \If {Path $\neq \emptyset$}
            \State $\robotcommandvel \gets$ FollowPath(Path)
            \If {PathFinished(Path)}
                \State Path $\gets \emptyset$
            \EndIf
        \Else
            \State $\robotcommandvel \gets$ LocalController(\parbox[t]{2.5in}{$\robotconfig[t], \deformconfig_t, \deformtarget,$ \\
                                                                              $\relaxeddistancematrix, \maxstretchfactor, \stretchingcorrectionweightfactor)$}
        \EndIf
        
        \State CommandConfiguration$(\robotconfig[t] + \robotcommandvel)$
        \State $\robotconfig[t+1] \gets$ SenseRobotConfig$()$
        \State $\deformconfig_{t+1} \gets$ SensePoints$()$
        \State $t \gets t + 1$
    \EndWhile
    \State \Return Success
\end{algorithmic}
\label{alg:mainloop}
\end{algorithm}

$\cspace_r$ and $\cspace_f$ are imbued with distance metrics $d_r(\cdot,\cdot) : \cspace_r \times \cspace_r \rightarrow \reals^{\geq 0}$ and $d_f(\cdot,\cdot) : (\cspace_r \times \bandspace) \times (\cspace_r \times \bandspace) \rightarrow \reals^{\geq 0}$, respectively. We define distances in robot configuration space and band space to be additive. I.e.
\begin{equation}
    d_f(\cdot, \cdot)^2 = d_r(\cdot, \cdot)^2 + \banddistscale d_b(\cdot, \cdot)^2
    \label{eqn:dist_metric}
\end{equation}
for some scaling factor $\banddistscale > 0$. To measure distances in $\bandspace$, we first upsample each band using linear interpolation to use the maximum number of points $\maxbandpoints$ for the given task, then measure the Euclidean distance between the upsampled points when considered as a single vector (Alg.~\ref{alg:band_dist}).

For a given planning problem, we are given a query $( \qinit_f, \Qgoal_f )$ which describes the initial configuration of the robot and band, as well as a goal region for the system to reach.  Note that $\Qgoal_f$ is defined implicitly via the GoalCheck() function and the parameters $(\eepositiongoal, \goalreachradius,\textrm{ Blacklist})$ rather than any explicit enumeration.

\begin{algorithm}[t]
\caption{BandDistance: $d_b(\band_1, \band_2)$}
\begin{algorithmic}[1]
    \State $\tilde \band_1 \gets$ UpsamplePoints$(\band_1, \maxbandpoints)$
    \State $\tilde \band_2 \gets$ UpsamplePoints$(\band_2, \maxbandpoints)$
    \State \Return $\| \tilde \band_1 - \tilde \band_2\|$
\end{algorithmic}
\label{alg:band_dist}
\end{algorithm}


We now establish a relationship between a path in robot configuration space $\cspacepath_r$ and one in the full configuration space $\cspacepath_f$ by making the following assumption.

\begin{assumption}[Deterministic Propagation]
\label{ass:deterministic}
    Given an initial configuration in full space $\qinit_f \in \cfree_f$ and the corresponding robot configuration $\qinit_r \in \cfree_r$, a path $\cspacepath_r : [0, 1] \rightarrow \cfree_r$ in robot configuration space with $\cspacepath_r(0) = \qinit_r$ uniquely defines a single path in full space $\cspacepath_f$, where $\cspacepath_f(0) = \qinit_f$.  Specifically, define
    \begin{multline}
        \cspacepath_f(t) =\\ \begin{bmatrix} \cspacepath_r(t) \\ 
                \lim_{h \rightarrow 0^-} \textup{ForwardPropogateBand}(\band(t-h), \cspacepath_r(t)) \end{bmatrix} \enspace .
        \label{eqn:deterministic}
    \end{multline}
\end{assumption}

Eq.~\eqref{eqn:deterministic} implicitly defines an underactuated system where the only way we can change the state of the band is by moving the robot; for a path in the full configuration space $\cspacepath_f$ to be achievable there must be a robot configuration space path $\cspacepath_r$, which when propagated using Eq.~\eqref{eqn:deterministic}, produces $\cspacepath_f$. Let $\textrm{FullSpace}(\cspacepath_r, \qinit_f)$ be the function that maps a given robot configuration space path $\cspacepath_r$ and full space initial configuration $\qinit_f$ to the full space path defined by Eq.~\eqref{eqn:deterministic}.

\subsection{Planning Problem Statement}

For a given planning instance, the task is to find a path starting from $\qinit_f$ through $\cfree_f$ to any point in $\Qgoal_f$, while obeying the constraints implied by Eq.~\eqref{eqn:deterministic}.

For a sequence of robot configurations $\qinit_f, \config_{1,r}, \dots, \config_{M,r} \in \cspace_r$, let $\cspacepath_r = \textrm{Path}(\qinit_r, \config_{1,r}, \dots, \config_{M,r})$ be the path defined by linearly interpolating between each point in order. Then, formally, the problem our planner addresses is the following:
\begin{equation}
    \begin{aligned}
        & \text{find}   & & \{ \config_{1,r}, \dots, \config_{M,r} \} \\
        & \text{s.t.}   & & \cspacepath_r = \textrm{Path}(\qinit_r, \config_{1,r}, \dots, \config_{M,r}) \\
        &               & & \cspacepath_f(s) \in \cfree_f, \; \forall s \in [0, 1] \\
        &               & & \cspacepath_f(1) \in \Qgoal_f \enspace .
    \end{aligned}
    \label{eqn:planning_problemstatement}
\end{equation}
\noindent where $\cspacepath_f = \textrm{FullSpace}(\cspacepath_r, \qinit_f)$.

\subsection{RRT-EB}

\begin{algorithm}[t]
\caption{RRT-EB$($\parbox[t]{2in}{$\robotconfig[t], \band_t, \eepositiongoal, \goalreachradius, \bandgoal,$ \\ $\maxbandlength, \bestneardist, \goalbias)$}}
\begin{algorithmic}[1]
    \State $\rrtnodeset \gets \{(\robotconfig[t], \band_t)\}$
    \State $\rrtedgeset \gets \emptyset$
    \State $\Qapproxgoal \gets$ GetGoalConfigs$(\eepositiongoal)$
    
    \While {$\neg$MaxTimeEllapsed()}
            \State $\qrand_f \gets$ SampleUniformConfig() \label{alg:bandrrt:basic_start}
            \State $\qnear_f \gets$ BestNearest$(\rrtnodeset, \rrtedgeset, \bestneardist, \qrand_f)$
            \State $\nodesnew, \edgesnew \gets$ Connect$(\qnear_f, \qrand_r, \maxbandlength)$
            \State $\rrtnodeset \gets \rrtnodeset \cup \nodesnew$
            \State $\rrtedgeset \gets \rrtedgeset \cup \edgesnew$
            
            \If {GoalCheck$(\nodesnew, \eepositiongoal, \goalreachradius, \bandgoal) = 1$}
                \State \Return ExtractPath$(\rrtnodeset, \rrtedgeset)$
            \EndIf \label{alg:bandrrt:basic_end}
            
            \State $\gamma \sim$ Uniform$[0, 1]$ \label{alg:bandrrt:bias_start}
            \If {$\gamma \leq \goalbias$}
                \State $\qlast_f \gets$ LastConfig$(\qnear_f, \nodesnew)$ \label{alg:bandrrt:lastconfig}
                \State $\qbias \gets \argmin_{\robotconfig \in \Qapproxgoal} d_r(\qlast_r, \robotconfig)$
                \State $\nodesnew, \edgesnew \gets$ Connect$(\qlast_f, \qbias, \maxbandlength)$
            \State $\rrtnodeset \gets \rrtnodeset \cup \nodesnew$
            \State $\rrtedgeset \gets \rrtedgeset \cup \edgesnew$
                
                \If {GoalCheck$(\nodesnew, \eepositiongoal, \goalreachradius, \bandgoal) = 1$}
                    \State \Return ExtractPath$(\rrtnodeset, \rrtedgeset)$
                \EndIf
            \EndIf \label{alg:bandrrt:bias_end}
    \EndWhile
    \State Return Failure
\end{algorithmic}
\label{alg:bandrrt}
\end{algorithm}

\begin{algorithm}[t]
\caption{BestNearest$(\rrtnodeset, \rrtedgeset, \bestneardist, \qrand_f)$}
\begin{algorithmic}[1]
    \State $\Qnear_f \gets \{ \config_f | \config_f \in \rrtnodeset, d_f(\config, \qrand_f) \leq \bestneardist \}$
    \If {$\Qnear_f \neq \emptyset$}
        \State \Return $\argmin_{\rrtnode \in \rrtnodeset} \cost(\rrtnode, \rrtnodeset, \rrtedgeset)$
    \Else
        \State $\Dnear[r]^2 \gets \min_{\config \in \rrtnodeset}{d_r(\qrand, \robotconfig)^2}$ \label{alg:nearst:robotspace}
        \State $\Dmax[f]^2 \gets \Dnear[r]^2\ + \banddistscale \Dmax[b]^2$
        \State $\Qnear_f \gets \{ \config | \config \in \rrtnodeset, d_r(\robotconfig, \qrand_r)^2 \leq \Dmax[f]^2 \}$ \label{alg:nearest:radius}
        \State \Return $\argmin_{\config_f \in \Qnear_f}{d_f(\config_f, \qrand_f)}$ \label{alg:nearest:fullspace}
    \EndIf
\end{algorithmic}
\label{alg:nearest}
\end{algorithm}

Our planner, RRT for Elastic Bands (RRT-EB), (Alg.~\ref{alg:bandrrt}) is based on an RRT with changes to account for a virtual elastic band in addition to the robot configuration. Lines \ref{alg:bandrrt:basic_start}-\ref{alg:bandrrt:basic_end} perform random exploration with lines \ref{alg:bandrrt:bias_start}-\ref{alg:bandrrt:bias_end} biasing the tree expansion towards the goal region. The key variations are the BestNearest function (Alg.~\ref{alg:nearest}) and the goal bias method.

\begin{figure}
    \centering
    \includegraphics[width=\columnwidth]{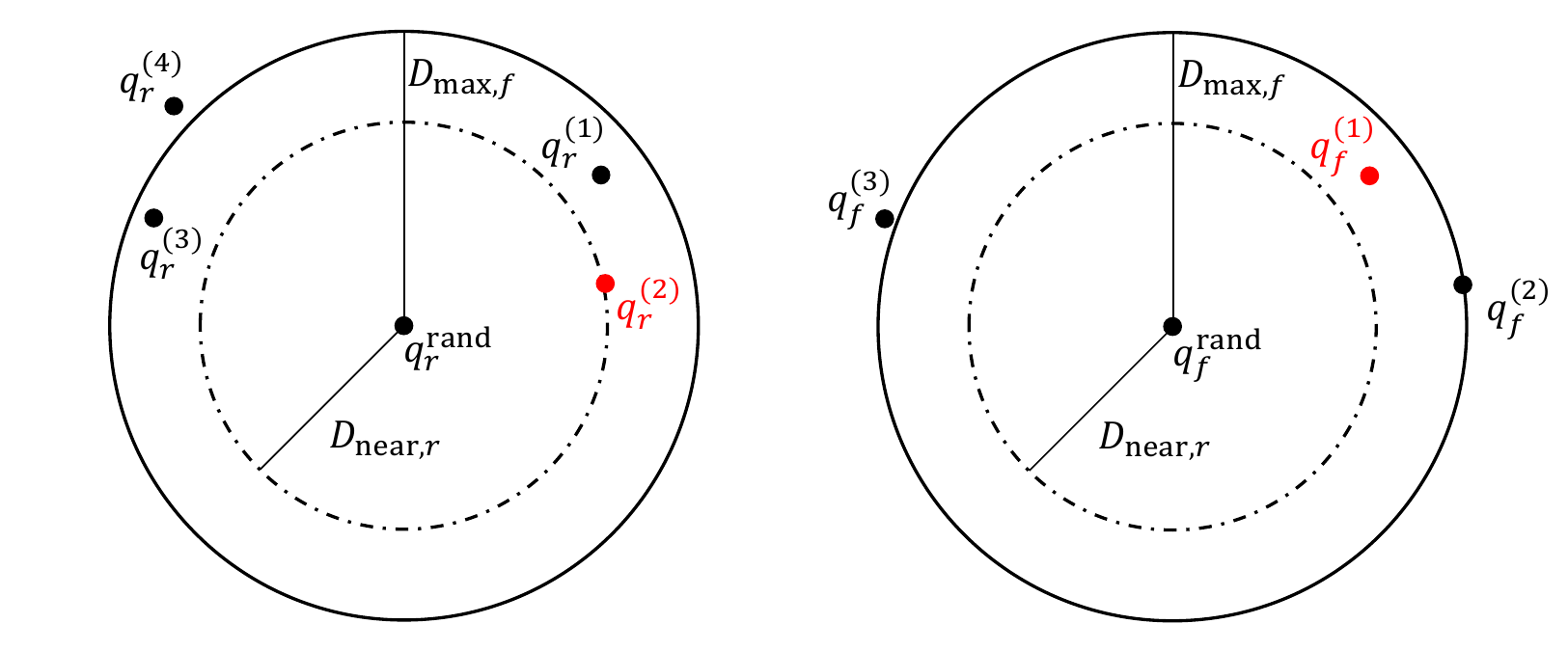}
    \caption{Left: $\config_r^{(2)}$ is the nearest node to the $\qrand$ in robot space, but it my be as far as $D_{\textrm{max}, f}$ away in the full configuration space. By considering all nodes within $D_{\textrm{max}, f}$ in robot space, we ensure that any node (such as $\config_f^{(1)}$) that is closer to $\qrand_f$ than $\config_f^{(2)}$ is selected as part of $\Qnear_f$, while nodes such as $\config_f^{(4)}$ are excluded in order to avoid the expense of calculating the full configuration space distance. Right: we then measure the distance in the full configuration space to all nodes that could possibly be the nearest to $\qrand_f$, returning $\config_f^{(1)}$ as the nearest node in the tree.}
    \label{fig:nearest}
\end{figure}

BestNearest is based on the selection method used by~\cite{LiAOKP2016}, selecting the node of smallest cost within a radius $\bestneardist$ if one exists, falling back to standard nearest neighbour behaviour if no node in the tree is within $\bestneardist$ of the random sample. We use path length in robot configuration space $\cspace_r$ as a cost function in our implementation. This helps reduce path length and ensures that we can specify lower bounds in Sec.~\ref{sec:delta_sim_traj_construction}. In order to avoid calculating distances in the full configuration space when it is not necessary, our method for finding the nearest neighbor is split into two parts, first searching in robot space, then searching in the full configuration space (see Fig.~\ref{fig:nearest}). Sec.~\ref{sec:nn_equiv} shows that this method is equivalent to searching in the full configuration space directly. $\bestneardist$ is an additional parameter compared to a standard RRT; it controls how much focus is placed on path cost versus exploration. The smaller $\bestneardist$, the less impact it has as compared to a standard RRT.  The larger $\bestneardist$ is, the harder it is to find narrow passages. We discuss further constraints on $\bestneardist$ in Section \ref{sec:select}.

To sample $\qrand_f = (\qrand_r,B^\textrm{rand})$, we sample the robot and band configurations independently, then combine the samples. For typical robot arms $\qrand_r$ is generated by sampling each joint independently and uniformly from the joint limits. To sample from $\bandspace$, we draw a sequence of $\maxbandpoints$ points from the bounded workspace. For our example tasks, workspace is a rectangular prism, and we sample each axis independently and uniformly.

Due to the fact that our system is highly underactuated, and the goal region is defined implicitly by a function call rather than an explicit set of configurations, we cannot sample from the goal set directly as is typically done for a goal bias. Instead we precompute a finite set of robot configurations $\Qapproxgoal$ such that the end-effectors of the robot are at $\eepositiongoal$. Then, as a goal bias mechanism, $\goalbias$ percent of the time, we attempt to connect to a potential goal configuration starting from the last configuration created by a call to the Connect function (or the last node selected by BestNearest if $\nodesnew = \emptyset$). A connection is then attempted between $\qlast_f$ and the nearest configuration in $\Qapproxgoal$. This allows us to bias exploration toward the robot component of the goal region, which we are able to define explicitly.


\section{Probabilistic Completeness of Global Planning}
\label{sec:analysis}

Proving probablistic completeness in $\cspace_f$ is challenging due to the multi-modal nature of the problem. Specifically, as the virtual elastic band moves in and out of contact the dimensionality of the manifold that the system is operating in can change. In addition, the virtual elastic band forward propagation function (Alg.~\ref{alg:band_propogation}) can allow the band to ``snap tight'' as the grippers move past the edge of an obstacle, changing the number of points in the band representation as it does so. By leveraging the assumptions from Sec.~\ref{sec:rpath_assumptions}, we are able to bypass most of these challenges by focusing on the portion of $\cspace_f$ that can be analyzed; i.e. $\cspace_r$.

This section proves the probabilistic completeness of the planning approach in two major steps.  First, it will show that the approach for selecting the nearest node in the tree for expansion is equivalent to performing a nearest-neighbor query in the full space.  Second, it proves that our algorithm will eventually return a path that is $\delta_r$-similar to an optimal $\delta$-robust solution to the planning problem with probability 1 (if it exists), or it will terminate early having found an alternate path to the goal region. Recall that we do not require an optimal path, only a feasible one.

\subsubsection{Assumptions and Definitions:}
\label{sec:rpath_assumptions}
Our problem allows for the virtual elastic band to be in contact with the surface of an obstacle, both during execution and as part of the goal set; this means that common assumptions regarding the expansiveness (\cite{Hsu1999}) of the planning problem may not hold. Instead of relying on expansiveness, we will define a series of alternate definitions and assumptions which are sufficient to ensure the completeness of our method.

First, in line with prior work, we will be assuming properties of the problem instance in regards to robustness.  In particular, we will be assuming the existence of a solution to a given query $\rpath_f : [0, 1] \rightarrow \cfree_f$ which has several robustness properties.  This solution is called a reference path.

To begin describing the properties of the reference path, we assume $\rpath_f$ has robustness properties in the robot configuration space.  That is, the corresponding path in robot configuration space $\rpath_r$ has strong $\delta_r$-clearance under distance metric $d_r(\cdot,\cdot)$ for some $\delta_r > 0$.

\begin{definition}[Strong $\delta$-clearance]
    A path $\cspacepath : [0, 1] \rightarrow \cfree$ has strong $\delta$-clearance under distance metric $d(\cdot,\cdot)$ if $\forall s \in [0, 1],\; d(\cspacepath(s),\cinv) \geq \delta$, for $\delta > 0$.
\end{definition}


Given our assumption about the $\delta_r$-clearance of the reference path in robot space, there exists a set $\setofpaths_r$ of $\delta_r$-similar paths to the reference path which are also collision-free.

\begin{definition}[$\delta$-similar path]
    Two paths $\cspacepath_a$ and $\cspacepath_b$ are $\delta$-similar if the Fr\'echet distance between the paths is less than or equal to $\delta$.
\end{definition}

Informally the Fr\'echet distance is described as follows (\cite{Alt1995Frechet}): Suppose a man is walking a dog. The man is walking on one curve while the dog on another curve. Both walk at any speed but are not allowed to move backwards. The Fr\'echet distance of the two curves is then the minimum length of leash necessary to connect the man and the dog.

Given the relationship between robot-space and full-space paths, we can define a full-space equivalent to $\setofpaths_r$ as
\begin{equation}
\begin{split}
    \setofpaths_f = \{ \cspacepath_f \mid \enspace &\cspacepath_r \in \setofpaths_r \textrm{ and} \\
                                                   &\cspacepath_f = \textrm{FullSpace}(\cspacepath_r, \qinit_f) \} \enspace .
\end{split}
\end{equation}





Given these assumptions and definitions, we are ready to define an \textit{acceptable $\delta$-robust path}:

\begin{definition}[Acceptable $\delta$-Robust Path]
\label{def:robust}
A path $\rpath_f$ is acceptable $\delta$-robust if the following hold:
\begin{enumerate}
    \item The robot-space reference path $\rpath_r$ has strong $\delta_r$-clearance for some $\delta_r > 0$;
    \item The final state for every path $\cspacepath_f \in \setofpaths_f$ is in $\Qgoal_f$. 
\end{enumerate}
\end{definition}

\noindent We assume there exists a reference path which satisfies this property and answers our given planning query:

\begin{assumption}[Solvable Problem]
    There exists some $\delta_r > 0$ such that the planning problem admits an acceptable $\delta$-robust path.
    \label{ass:solvable_problem}
\end{assumption}

If a planning problem does not yield a reference path with this property, then it would be practically impossible for a sampling-based approach to solve it, as this would require sampling on a lower-dimensional manifold in robot space. Given that our planner is able to find paths, we believe this assumption is true except in special cases where the band must achieve a singular configuration to reach the goal.

While the focus of this paper is not on asymptotic optimality, we will make use of a cost function $\cost(\cspacepath)$ of a path in Sec.~\ref{sec:select}. Our cost function is path length in robot configuration space. With a cost function of this form we then assume from here onward that the reference path in question is optimal under the following definition.

\begin{definition}[Optimal $\delta$-Robust Path]
    Let $\setofpaths_{f,\delta}$ be the set of all acceptable $\delta$-robust paths. A path $\rpath_f$ is optimal $\delta$-robust if
    \begin{equation}
        \cost(\rpath_f) = \inf_{\cspacepath_f \in \setofpaths_{f,\delta}} \cost(\cspacepath_f) \enspace .
    \end{equation}
\end{definition}

Finally, we also assume that workspace is bounded. This will be true for any practical task and is rarely mentioned in the literature, but we will use this assumption in our analysis in Sec.~\ref{sec:nn_equiv}.

\subsection{Proof of Nearest-Neighbors Equivalence}
\label{sec:nn_equiv}

\begin{lemma}
    \label{lem:banddist}
     If the maximum distance between any two points in workspace is bounded by $\Dmax[w] > 0$, then under distance metric $d_b(\cdot, \cdot)$, the maximum distance between any two points in virtual elastic band space is bounded. I.e. $\exists \Dmax[b] > 0$ such that $d_b(\band_1, \band_2) \leq \Dmax[b] \; \forall \band_1, \band_2 \in \bandspace$.
\end{lemma}

\noindent
{\bf Proof.}
From the definition of $\bandspace$ in Sec.~\ref{sec:overstretch}, the number of points used to represent a virtual elastic band is bounded by $\maxbandpoints$. Let $\band_1, \band_2 \in \bandspace$ be two virtual elastic band configurations, and let $\tilde \band_1 = (\tilde b_{1,1}, \dots, b_{1,\maxbandpoints})$ and $\tilde \band_2 = (\tilde b_{2,1}, \dots, b_{2,\maxbandpoints})$ be their upsampled versions as described in Alg.~\ref{alg:band_dist}. Then
\begin{equation}
\begin{split}
    d_b(\band_1, \band_2)^2 &= \sum_{i=1}^{\maxbandpoints} \| \tilde b_{1,i} - \tilde b_{2,i} \|^2 \\
                            &\leq \sum_{i=1}^{\maxbandpoints} \Dmax[w]^2 = \maxbandpoints \Dmax[w]^2 = \Dmax[b]^2
\end{split}
\end{equation}
\qed

\begin{lemma}
    If workspace is bounded, then lines \ref{alg:nearst:robotspace}-\ref{alg:nearest:fullspace} in Alg.~\ref{alg:nearest} are equivalent to a nearest neighbor search in the full configuration space directly.
\end{lemma}

\noindent
{\bf Proof.}
The upper bound of $\Dmax[b]$ and our additive distance metric (Eq.~\eqref{eqn:dist_metric}) ensures that the distance between any two configurations in full space $\cspace_f$ can be bounded using only the distance in robot configuration space:
\begin{equation}
    d_f(\cdot,\cdot)^2 \leq d_r(\cdot,\cdot)^2 + \banddistscale \cdot \Dmax[b]^2 \enspace .
    \label{eqn:bounded_full_distance}
\end{equation}
Next, consider that in Line~\ref{alg:nearst:robotspace} of the algorithm, the nearest neighbor to $\qrand_r$ under distance metric $d_r$ is found.  Let this nearest neighbor be denoted $\qnearapprox_r$, keeping in mind that it belongs to a vertex in the tree $\qnearapprox_f = (\qnearapprox_r, \qnearapprox_b)$.  Let the (squared) distance between these points under $d_r$ be $\Dnear[r]^2$.  From Eq.~\eqref{eqn:bounded_full_distance}, we can bound the distance between the random sample and $\qnearapprox_f$ under $d_f$ as $\Dmax[f]^2 \leq \Dnear[r]^2 + \banddistscale \Dmax[b]^2 = \Dmax[f]^2$.

In Line~\ref{alg:nearest:radius} of the algorithm, a radius nearest-neighbors query of radius $\Dmax[f]$ is performed, returning a set $\Qnear_f$.  By construction if there is a node $\config_f \in \rrtnodeset$ that is closer to $\qrand_f$ than $\qnearapprox_f$, then $\config_f \in \Qnear_f$ (Fig.~\ref{fig:nearest}). Then, the method selects as the true nearest neighbor in full space $\config^\textrm{select}_f = \argmin_{\config_f \in \Qnear_f}{d_f(\config_f, \qrand_f)}$.
\qed



\begin{figure}
    \centering
    \includegraphics[width=\columnwidth]{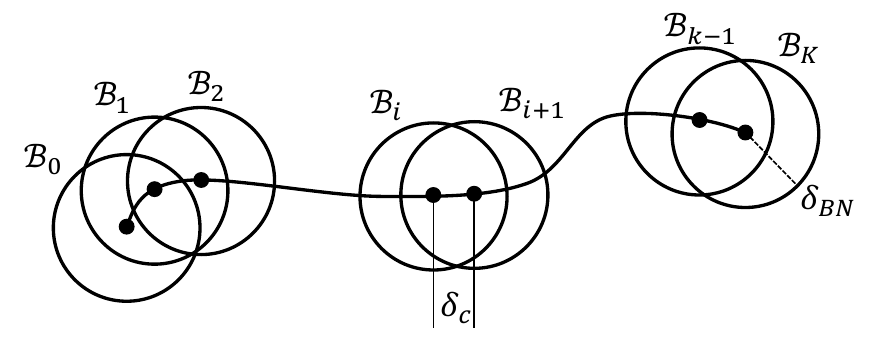}
    \caption{Example covering ball sequence for an example reference path with a distance along the path of $\ballseparation$ between each ball. Given that the path is $\delta_r$-robust, each ball is a subset of $\cfree_r$.}
    \label{fig:covering_ball_sequence}
\end{figure}

\subsection{Construction of a $\delta_r$-similar Path}
\label{sec:delta_sim_traj_construction}

The objective here is to show with probability approaching $1$, the planner generates a $\delta_r$-similar path to some robustly-feasible solution given enough time.  If an alternate path is found and the algorithm terminates before generating a $\delta_r$-similar path then this is still sufficient for probabilistic completeness.  This analysis is similar to \cite{LiAOKP2016}, and is based on a covering ball sequence of the optimal  $\delta$-robust path $\rpath_r$. \rev{The key differences are in Sec.~\ref{sec:prop} where we show that using a straight line to connect points in $\cspace_r$ is sufficient to get a lower bound on the probability of covering the next ball, while Li \textit{et. al.} used a random control action.}

\begin{definition}[Covering Ball Sequence]
    \label{def:coveringballseq}
    Given a path $\cspacepath_r : [0, 1] \rightarrow \cfree_r$, robust clearance $\delta_r > 0$, a BestNearest distance $\bestneardist > 0$, and a distance value $0 < \ballseparation < \bestneardist < \delta_r$; the covering ball sequence is defined as a set of $K + 1$ hyper-balls $\{ \hyperball_{\delta_r}(\config_{0,r}), \dots, \hyperball_{\delta_r}(\config_{K,r}) \}$ of radius $\delta_r$, where $\config_{k,r}$ are defined such that:
    \begin{itemize}
        \item $\config_{0,r} = \cspacepath_r(0)$;
        \item $\config_{K,r} = \cspacepath_r(1)$;
        \item $\textup{PathLength}(\config_{k-1,r}, \config_{k,r}) = \ballseparation$ for $k = 1, \dots, K$.
    \end{itemize}
\end{definition}
\noindent
Denote $\config^*_{k,r}$ to be the center of the $k^{th}$ covering hyper-ball for the reference path $\rpath_r$. Fig.~\ref{fig:covering_ball_sequence} shows an example of a covering ball sequence.

The objective is to show that the vertex set of the planning tree after $n$ iterations $\rrtnodeset_n$ probabilistically contains a node within the goal set, i.e.
\begin{equation}
    \liminf_{n \rightarrow \infty} \pr( \rrtnodeset_n \cap \Qgoal_f \neq \emptyset ) = 1 \enspace .
\end{equation}
To do this, the analysis examines $K$ subsegments of the reference path $\rpath_r$, based on the covering ball sequence for the reference path. If we can generate a robot path that is $\delta_r$ similar to $\rpath_r$, then given Assumption~\ref{ass:solvable_problem} and the properties of the reference path, the corresponding full space path will be a solution to the given planning problem.

Let $A_k^{(n)}$ be the event that on the $n^{th}$ iteration of the algorithm, it generates a $\delta_r$-similar path to the $k^{th}$ subsegment of $\rpath_r$.  This of course requires two events to occur: the node generated from the prior propagation covering segment $k-1$ must be selected for expansion, and the expansion must then produce a $\delta_r$-similar path to the current segment.  Then, let $E_k^{(n)}$ be the event that for segment $k$, $A_k^{(n)}$ has occurred for some $i \in [1,n]$, i.e. $E_k^{(n)}$ indicates whether the algorithm has constructed the $\delta_r$-similar edge for subsegment $k$. From these definitions, the goal then is to show that
\begin{equation}
    \lim_{n \rightarrow \infty} \pr(\textrm{Success}) = \lim_{n \rightarrow \infty} \pr\left( E^{(n)}_K \right) = 1 \enspace.
    \label{eqn:initial_pr_limit}
\end{equation}

We start by considering the probability of failing to generate an arbitrary segment $1 \leq k \leq K$. Then 
\begin{equation}
\begin{split}
    \pr&\left( \neg E^{(n)}_k \right) \\
       &= \pr\left( \neg A^{(1)}_k \cap \dots \cap \neg A^{(n)}_k \right) \\
       &= \pr\left( \neg A^{(1)}_k \right) \pr\left( \neg A^{(2)}_k \mid \neg A^{(1)}_k \right) \cdot \dots \\
       &\hspace{1.5cm} \cdot \pr\left( \neg A^{(n)}_k \mid \neg A^{(1)}_k \cap \dots \cap \neg A^{(n-1)}_k \right) \\
       &= \prod_{i=1}^n \pr\left( \neg A^{(i)}_k \mid \neg E^{(i-1)}_k \right) \enspace .
\end{split}
\label{eqn:fail_enk}
\end{equation}
Note the definition of $\neg E^{(i-1)}_k$ is what allows us to collapse the product into a concise form.

The probability that $\neg A^{(i)}_k$ happens given $\neg E^{(i-1)}_k$ is equivalent to the probability that we have not yet generated a $\delta_r$-similar path for segment $k-1$ (i.e. $\pr( \neg E^{(i-1)}_{k-1} )$) plus the probability that the previous segment has been generated, but we fail to generate the current segment:
\begin{equation}
\begin{split}
    \pr&\left( \neg A^{(i)}_k \mid \neg E^{(i-1)}_k \right) \\
    &= \pr\left( \neg E^{(i-1)}_{k-1} \right) + \pr\left( E^{(i-1)}_{k-1} \right) \\
    &  \hspace{1.7cm} \cdot \pr\left( \neg A_k^{(i)} \mid E_{k-1}^{(i-1)} \cap \neg E_k^{(i-1)} \right), \\
\end{split}
\end{equation}
which we can rewrite in terms of $A_k^{(i)}$ instead of $\neg A_k^{(i)}$:
\begin{equation}
\begin{split}
    \pr&\left( \neg A^{(i)}_k \mid \neg E^{(i-1)}_k \right) \\
       &= \pr\left( \neg E^{(i-1)}_{k-1} \right) + \pr\left( E^{(i-1)}_{k-1} \right) \\
       &  \hspace{1.7cm} \cdot \left(1 - \pr\left( A_k^{(i)} \mid E_{k-1}^{(i-1)} \cap \neg E_k^{(i-1)} \right) \right) \enspace . \\
\end{split}
\end{equation}
Then multiplying out the last term we get
\begin{equation}
\begin{split}
    \pr&\left( \neg A^{(i)}_k \mid \neg E^{(i-1)}_k \right) \\
       &= \pr\left( \neg E^{(i-1)}_{k-1} \right) + \pr\left( E^{(i-1)}_{k-1} \right) \\
       &  \hspace{0.65cm} - \pr\left( E^{(i-1)}_{k-1} \right) \pr\left( A_k^{(i)} \mid E_{k-1}^{(i-1)} \cap \neg E_k^{(i-1)} \right) \enspace . \\
\end{split}
\end{equation}
Finally, summing the first two terms, we arrive at
\begin{equation}
\begin{split}
    \pr&\left( \neg A^{(i)}_k \mid \neg E^{(i-1)}_k \right) \\
       &= 1 - \pr\left( E^{(i-1)}_{k-1} \right) \pr\left( A_k^{(i)} \mid E_{k-1}^{(i-1)} \cap \neg E_k^{(i-1)} \right) \enspace .
\end{split}
\end{equation}
Two events need to happen in order to generate a path to the next hyperball; an appropriate node must be selected for expansion, and Connect$(\dots)$ must generate a $\delta_r$-similar path segment, assuming that the appropriate node has already been selected. Denote the probability of these events at iteration $i$ as $\selectprobability$ and $\propagationprobability$ respectively. Then
\begin{equation}
    \pr\left( \neg A^{(i)}_k \mid \neg E^{(i-1)}_k \right) = 1 - \pr\left( E^{(i-1)}_{k-1} \right) \selectprobability \propagationprobability \enspace .
    \label{eqn:fail_aik}
\end{equation}

\noindent
As we are examining this probability in the limit, we will instead draw a bound on this probability to put it in a form we can easily examine the limit for. To do so, we must carefully consider the values of $\selectprobability$ and $\propagationprobability$.  In Section~\ref{sec:select}, it will be shown that $\selectprobability$ is a generally decreasing function, but converges to a finite value $\selectbound > 0$ in the limit.  Therefore we let $\selectbound$ be a lower bound of $\selectprobability$.  Then in Section~\ref{sec:prop}, $\propagationprobability$ will similarly be shown to be positive and lower-bounded; in particular $\selectprobability \propagationprobability \leq \selectbound$.  Taking $\selectbound$ as constant, we can bound Eq.~\eqref{eqn:fail_aik} as
\begin{equation}
    \pr\left( \neg A^{(i)}_k \mid \neg E^{(i-1)}_k \right) \leq 1 - \pr\left( E^{(i-1)}_{k-1} \right) \selectbound \enspace .
    \label{eqn:fail_aik_bound}
\end{equation}
Combining equations \eqref{eqn:fail_aik_bound} and \eqref{eqn:fail_enk} we have
\begin{equation}
    \pr\left( \neg E^{(n)}_k \right) \leq \prod_{i=1}^n \left(1 - \pr\left( E^{(i-1)}_{k-1} \right) \selectbound \right) \enspace .
\end{equation}
Denote $y^{(n)}_k = \prod_{i=1}^n \left(1 - \pr\left( E^{(i-1)}_{k-1} \right) \selectbound \right)$. Then
\begin{equation}
    \pr\left( \neg E^{(n)}_k \right) \leq y^{(n)}_k \enspace .
    \label{eqn:fail_enk_bound}
\end{equation}
We will show using induction over $k$, that Eq.~\eqref{eqn:fail_enk_bound} tends to 0 as $n \rightarrow \infty$, and thus $\lim_{n \rightarrow \infty} \pr(\textrm{Success}) = 1$\\

\noindent
\textbf{Base case $(k = 1)$:}\\
Note that $\pr(E^{(i)}_{0}) = 1$ because the start node always exists. Then 
\begin{equation}
\begin{split}
    \lim_{n \rightarrow \infty} \pr\left( \neg E^{(n)}_1 \right) 
        &\leq \lim_{n \rightarrow \infty} \prod_{i=1}^n \left(1 - \pr\left( E^{(i-1)}_{0} \right) \selectbound \right) \\
        &=    \lim_{n \rightarrow \infty} \prod_{i=1}^n \left(1 - \selectbound \right) \\
        &=\lim_{n \rightarrow \infty} \left(1 - \selectbound \right)^n = 0 \enspace .
\end{split}
\end{equation}

\noindent
\textbf{Induction hypothesis: }
\begin{equation}
    \lim_{n \rightarrow \infty} \pr\left( \neg E^{(n)}_m \right) = 0 \textrm{ for } m = 1, 2, \dots, k - 1 \enspace .
    \label{eqn:induction_hypothesis}
\end{equation}
Note that this implies $\lim_{n \rightarrow \infty} \pr( E^{(n)}_m ) = 1$ for $m = 1, 2, \dots, k - 1$.\\

\noindent
\textbf{Induction step ($2 \leq k \leq K$): }\\
Consider the log of the bound on $\pr\left( \neg E^{(n)}_k \right)$:
\begin{equation}
    \log y^{(n)}_k = \sum_{i=1}^n \log\left( 1 - \pr\left( E^{(i-1)}_{k-1} \right) \selectbound \right) \enspace .
    \label{eqn:logy}
\end{equation}
Denote $x = \pr\left( E^{(i-1)}_{k-1} \right) \selectbound$. Given that $0 \leq x < 1$, and writing the Taylor series expansion of $\log\left( 1 - x \right)$ centered at $x = 0$ we have
\begin{equation}
    \log \left( 1 - x \right) = - \sum_{m=1}^\infty \frac{x^m}{m} \enspace .
    \label{eqn:log1minusx}
\end{equation}
Substituting Eq.~\eqref{eqn:log1minusx} back into Eq.~\eqref{eqn:logy} we get
\begin{equation}
    \log y^{(n)}_k = - \sum_{i=1}^n \sum_{m=1}^\infty \frac{\left( \pr\left( E^{(i-1)}_{k-1} \right) \selectbound \right)^m}{m} \enspace .
\end{equation}
Dropping all but the first term in the infinite sum we get the bound
\begin{equation}
    \log y^{(n)}_k \leq -\sum_{i=1}^n \pr\left( E^{(i-1)}_{k-1} \right) \selectbound \enspace .
\end{equation}
Rearranging terms yields
\begin{equation}
    \log y^{(n)}_k \leq -\selectbound \sum_{i=1}^n \pr\left( E^{(i-1)}_{k-1} \right) \enspace .
\end{equation}
We now use the induction hypothesis. We know that $\pr( E^{(n)}_{k-1} ) \rightarrow 1$ as $n \rightarrow \infty$, thus $\sum_{i=1}^n \pr( E^{(i-1)}_{k-1} ) \rightarrow \infty$. Then
\begin{equation}
    \lim_{n \rightarrow \infty} \log y^{(n)}_k \leq -\selectbound \sum_{i=1}^n \lim_{n \rightarrow \infty} \pr\left( E^{(i-1)}_{k-1} \right) = -\infty \enspace .
    \label{eqn:logy_bound}
\end{equation}
Taking the log of Eq.~\eqref{eqn:fail_enk_bound} and combining with Eq.~\eqref{eqn:logy_bound} we get
\begin{equation}
    \lim_{n \rightarrow \infty} \log \pr\left( \neg E^{(n)}_k \right) \leq \lim_{n \rightarrow \infty} \log y^{(n)}_k = -\infty
\end{equation}
and therefore
\begin{equation}
    \lim_{n \rightarrow \infty} \pr\left( \neg E^{(n)}_k \right) = 0 ,
\end{equation}
which completes the induction step.

Thus, given that $\pr(\neg E^{(n)}_k) \rightarrow 0$ as $n \rightarrow \infty$ for any $1 \leq k \leq K$
\begin{equation}
    \lim_{n \rightarrow \infty} \pr(\textrm{Success}) = \lim_{n \rightarrow \infty} \left( 1 - \pr\left( \neg E^{(n)}_K \right) \right) = 1 \enspace.
\end{equation}

\subsubsection{Selection of an appropriate node ($\selectbound$):}
\label{sec:select}

First, we define the following restriction on the definition of $\bestneardist$: 

\begin{definition}[$\bestneardist$ Restriction]
\label{prop:bestnear_requirement}
    For a reference path $\rpath_r$ with robustness $\delta_r$, $\bestneardist$ is defined such that $\innerballsize = \delta_r - \bestneardist > 0$.
\end{definition}

The proof that $\selectbound > 0$ follows directly from the related work of \cite{LiAOKP2016} (proof of Lemma 23).  To summarize, due to best-nearest neighbors selection, there exists a positive-measure region around the minimum cost vertex $\qnear_f$ which observes the optimal reference path in which its cost dominates all other nearby nodes, and therefore, when $\qrand_f$ is drawn in this volume, $\qnear_f = (\qnear_r, \qnear_b)$ is guaranteed to be selected (Fig.~\ref{fig:Yanbo_lemma_23_figure}).  Since our approach follows an equivalent sampling and nearest neighbor method to \cite{LiAOKP2016} Alg.~6 (as shown in Sec.~\ref{sec:nn_equiv}), 
\begin{equation}
    \selectbound = \frac{\mu\left( \hyperball_{\delta_\theta}\big( \config_{k,f}^* \big) \cap \hyperball_{\delta_{BN}}\big( \qnear_f \big) \right)}{\mu\left( \cspace_f \right)} > 0
\end{equation}
follows directly.

To show that $\selectbound < 1$, we need only consider the case when there are at least 2 nodes in $\rrtnodeset$.

\subsubsection{$\delta_r$-similar Propagation ($\propagationprobability$):}
\label{sec:prop}
Given that our nearest neighbor method is non-standard, and operating in the full configuration space $\cspace_f$, we need to carefully consider how this affects the propagation probability $\propagationprobability$. Given the kinematic model of our robot system, it is straightforward to show that the system in robot space is Small-Time Locally Controllable (STLC), i.e. $\robotconfig$ can be instantaneously moved in any direction, barring the presence of obstacles or configuration space limits.  

Then, based on the construction of the covering ball sequence and the $\bestneardist$ restriction, the following lemma holds.

\begin{figure}
    \centering
    \includegraphics[width=0.5\columnwidth]{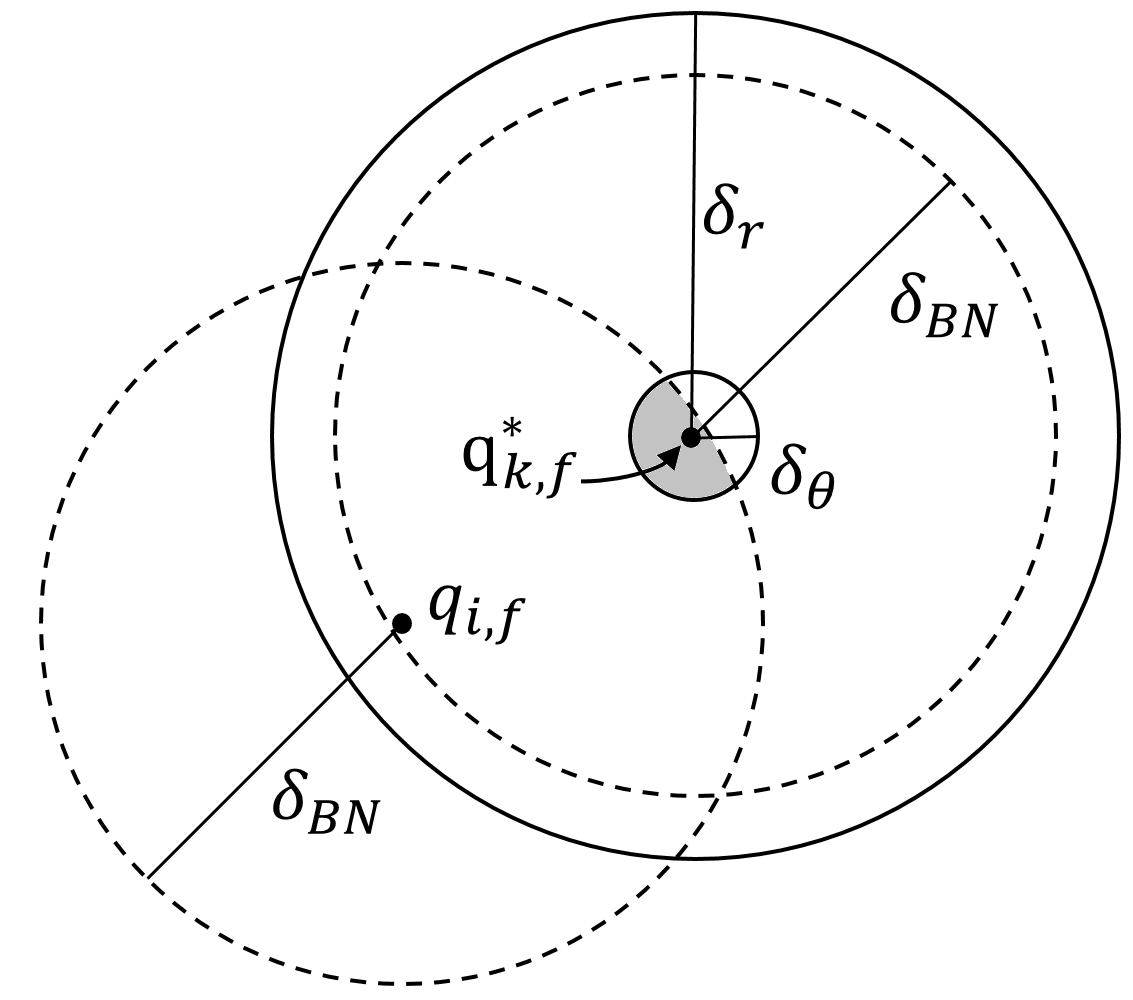}
    \caption{Minimum domination region for a node $\config_{i,f}$, adapted from \cite{LiAOKP2016} Lemma 23. Sampling $\qrand_f$ in the shaded region guarantees that a node $\qnear_f \in \hyperball_{\delta_r}(\config_{k,f}^*)$ is selected for propagation so that either $\qnear_f = \config_{i,f}$ or $\cost(\qnear_f) < \cost(\config_{i,f})$.}
    \label{fig:Yanbo_lemma_23_figure}
\end{figure}

\begin{lemma}
    \label{lem:rand_to_next_dist}
    If $\qrand_f$ is within the minimum domination region as described in \cite{LiAOKP2016} Lemma 23 (Fig.~\ref{fig:Yanbo_lemma_23_figure}), then $\qrand_r \in \hyperball_{\delta_r}(\config^*_{k, r})$ and Connect() will generate a segment that is $\delta_r$-similar to segment $k$ of the reference path.
\end{lemma}

\noindent
{\bf Proof.}
Assume that $\qrand_f \in \hyperball_{\innerballsize}(\config^*_{k-1,f})$. Then we have
\begin{align*}
    d_r(\qrand_r, \config^*_{k,r}) &\leq d_f(\qrand_f, \config^*_{k,f}) \\
                                   &\leq d_f(\qrand_f, \config^*_{k-1,f}) + d_f(\config^*_{k-1,f}, \config^*_{k,f}) \\
                                   &\leq \innerballsize + \ballseparation = \delta_r - \bestneardist + \ballseparation \enspace.
\end{align*}
Then by construction of the covering ball sequence, we have that $\ballseparation -\bestneardist < 0$ and thus $d_r(\qrand_r, \config^*_{k,r}) < \delta_r$. In addition, we have that the straight line between $\qnear_r$ as selected by $\qrand_r$ is entirely contained in $\hyperball_{\delta_r}(\config^*_{k-1,r})$, and thus is also in $\cfree_r$ as the reference path is optimal $\delta$-robust. We then have that the path generated by Connect is $\delta_r$-similar to the $k^{th}$ segment of the reference path.
\qed

\begin{lemma}
    The probability of covering segment $k$ at iteration $i$, given that we have not yet covered segment $k$ but we have covered segment $k-1$
    $$\pr\left( A_k^{(i)} \mid E_{k-1}^{(i-1)} \cap \neg E_k^{(i-1)} \right) = \selectprobability \propagationprobability$$
    is lower-bounded by $\selectbound$.
\end{lemma}

\noindent
{\bf Proof.}
Consider two possible events. First, that $\qrand_f$ is within the minimum domination region (Fig.~\ref{fig:Yanbo_lemma_23_figure}) of $\qnear_f$. If $\qrand_f$ is within the minimum domination region of $\qnear_f$, then by Lemma~\ref{lem:rand_to_next_dist}, Connect() will generate a $\delta_r$-similar segment with probability 1. Denote this event as $B$. Second, the event that $\qrand_f$ is somewhere else. Denote this event as $C$. Then we can bound $\pr( A_k^{(i)} \mid E_{k-1}^{(i-1)} \cap \neg E_k^{(i-1)})$ by considering only $B$:
\begin{align*}
    \pr\left( A_k^{(i)} \mid E_{k-1}^{(i-1)} \cap \neg E_k^{(i-1)} \right) 
        &= \pr(B) + \pr(C) \\
        &\geq \pr(B) \geq \selectbound \enspace .
\end{align*}
\qed

\section{Simulation Experiments and Results}
\label{sec:simulation_experiments}

We now present four example tasks to demonstrate our algorithm, two with cloth, and two with rope. These tasks are designed to show that our framework is able to handle non-trivial tasks which cannot be performed using either our controller or planner alone. In Sec.~\ref{sec:live_robot} we demonstrate that our method can also be applied to a physical robot.

For these simulation tasks $\cspace_r = \gripperconfigspace$ -- i.e. there are two free flying grippers. In the first and second tasks, two grippers manipulate the cloth so that it covers a table. In the first task the cloth is obstructed by a pillar while in the second task the grippers must pass through a narrow passage before the table can be covered. The third and fourth scenarios require the robot to navigate a rope through a three-dimensional maze before aligning the rope with a line traced on the floor (see Figure~\ref{fig:example_tasks}). The video accompanying this paper shows the task executions.

All experiments were conducted in the open-source Bullet simulator \citep{Coumans2010}, with additional wrapper code developed at UC Berkeley \citep{ucberkley_bullet}. The cloth is modeled as a triangle mesh using 1500 vertices with a total size of $0.3\text{m} \times 0.5\text{m}$. The rope is modeled as a series of small capsules linked together by springs. In the first rope experiment we use 39 capsules for a 0.78m long rope, and 47 capsules for a 0.94m rope in the last experiment. We emphasize that our method does not have access to the model of the deformable object or the simulation parameters. The simulator is used as a ``black box'' for testing. We set the maximum stretching factor $\maxstretchfactor$ to 1.17 for the cloth and 1.15 for the rope. All tests are performed using an i7-8700K 3.7 GHz CPU with 32 GB of RAM. We use the same deadlock prediction and planner parameters for all tasks, shown in Tables~\ref{tab:deadlock_param_table} and~\ref{tab:rrt_param_table}. For the purpose of the planner we treat the grippers as spheres, reducing the planning space from $\gripperconfigspace \times \bandspace$ to $\reals^6 \times \bandspace$. \rev{To lift the planned path back into $\gripperconfigspace \times \bandspace$ we copy the starting orientation of the grippers to each gripper configration in the plan.}

To smooth the path returned by the planner, at each iteration we randomly select either a single gripper or both grippers and two configurations in the path. To smooth between the configurations we use the same forward-propagation method for the virtual elastic band as used in the planning process. If we have selected only one gripper for smoothing, we do not change the configuration of the second gripper during that smoothing iteration. We also forward-propagate the virtual elastic band to the end of the path to ensure that the band at the end of the smoothed path is dissimilar from the blacklist. We perform 500 smoothing iterations for experiments 1, 2, and 4; and 1500 for experiment 3 due to the larger environment.

\subsection{Single Pillar}
\label{sec:single_pillar}

\begin{figure*}[t]
    \centering
    \includegraphics[width=\textwidth]{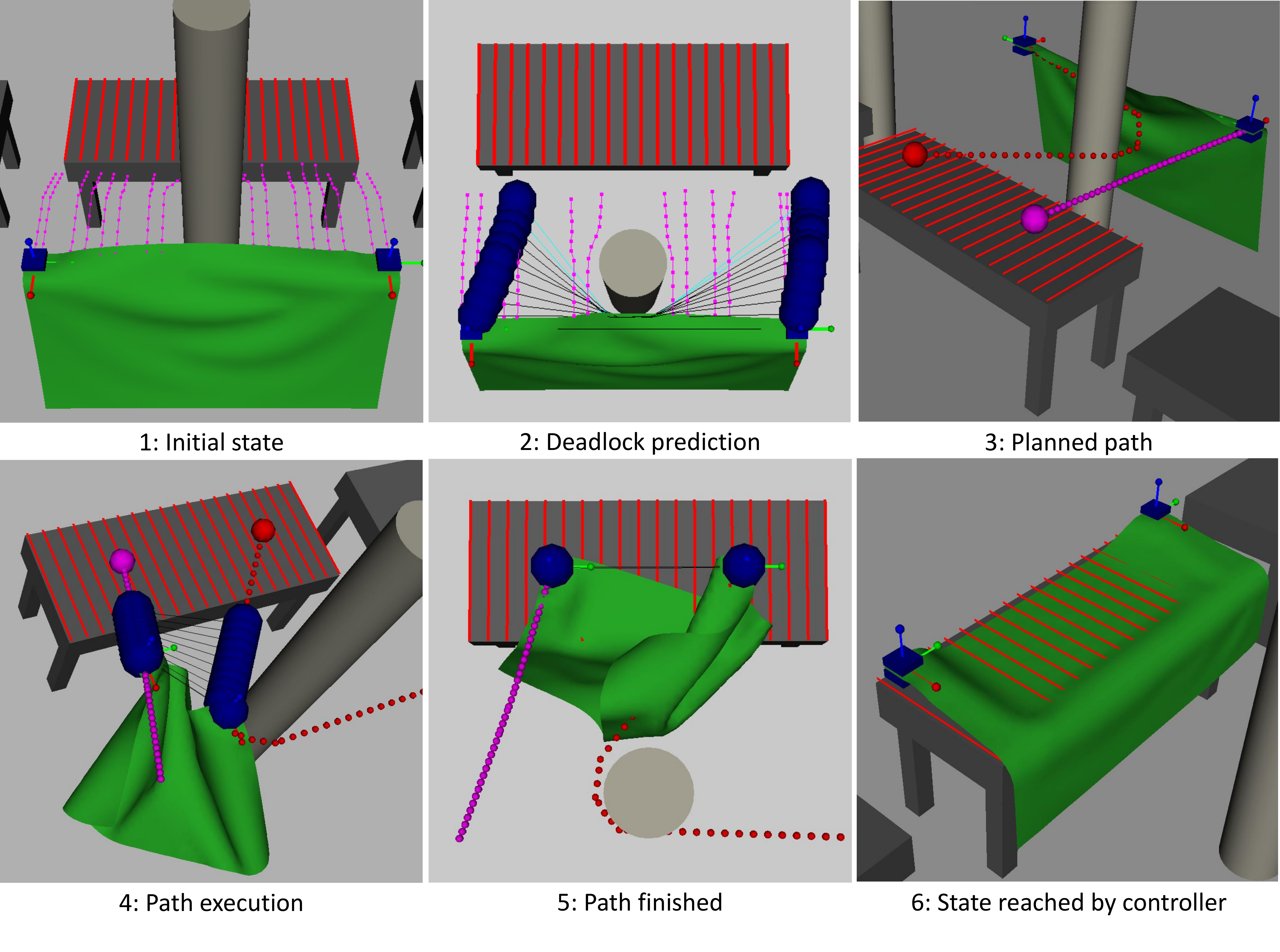}
    \vspace{-0.2in}
    \caption{Sequence of snapshots showing the execution of the first experiment. The cloth is shown in green, the grippers are shown in blue, and the target points are shown as red lines. (1) The approximate integration of the navigation functions from error reduction over $\predictionhorizon$ timesteps, shown in magenta, pull the cloth to opposite sides of the pillar. (2) A sequence of \textit{virtual elastic bands} between the grippers is shown in black and teal, indicating the predicted gripper configuration over the prediction horizon as the local controller follows the navigation functions. The elastic band changes to teal as the predicted motion of the grippers moves the cloth into an infeasible configuration. (3 - 5) The resulting plan by the RRT, shown in magenta and red, moves the system into a new neighbourhood. (6) Final system state when the task is finished by the local controller.}
    \label{fig:cloth_single_pole}
\end{figure*}

In the first example task, the objective is to spread the cloth across a table that is on the far side  of a pillar (see Figure~\ref{fig:cloth_single_pole}). We uniformly discretize the surface of the table to create the target points $\deformtarget$, with each discretized point creating a navigation function that pulls the closest point on the deformable object towards the target. These target points are set slightly above the surface to allow for collision margins within the simulator. A single point on the cloth can have multiple ``pulls'' or none. Task error $\errorfunction$ is defined as the sum of the Dijkstra's distances from each target point to the closest point on the cloth. If a target point in $\deformtarget$ is within a small-enough threshold of their nearest neighbors in $\deformconfig$, then these points are considered ``covered'' and do not influence task error or any other calculation. Our results show that even though the global planner is only planning using the gripper positions and a virtual elastic band between them, it is able to find the correct neighbourhood for the local controller to complete the task. On average we are able to find and smooth a path in 3.0 seconds (Table~\ref{tab:planning_statistics}), with the majority of the planning time spent on forward propagation of the virtual elastic band as part of the validity check for a potential movement of the grippers. In all 100 trials the global planner is only invoked once, with the local controller completing the task after the plan finishes.

\subsection{Double Slit}

\begin{figure*}[t]
    \centering
    \includegraphics[width=\textwidth]{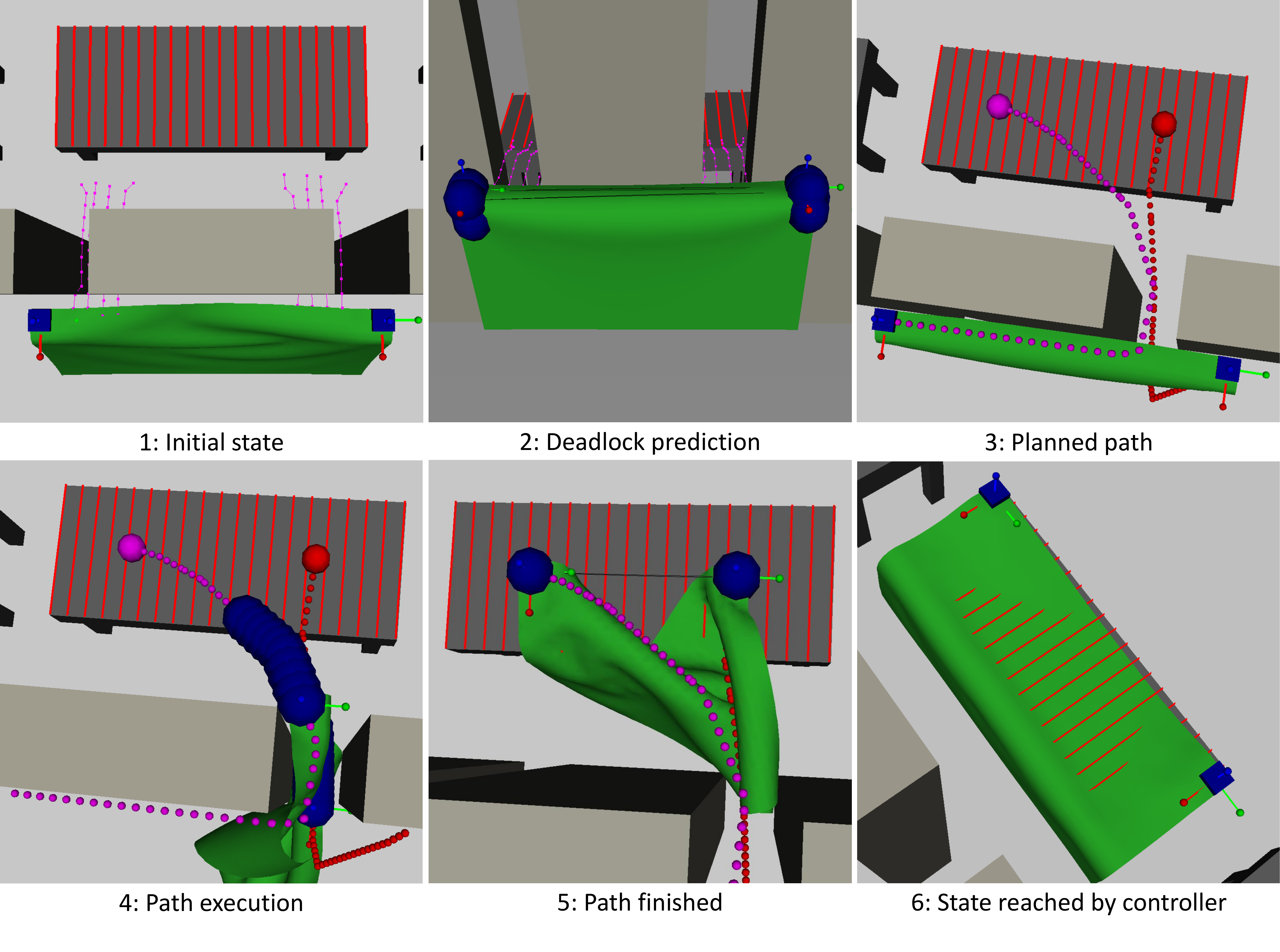}
    \vspace{-0.2in}
    \caption{Sequence of snapshots showing the execution of the second experiment. We use the same colors as the previous experiment (Figure~\ref{fig:cloth_single_pole}), but in this example instead of detecting future overstretch in panel (2), we detect that the system is stuck in a bad local minimum and unable to make progress.}
    \label{fig:cloth_double_slit}
\end{figure*}

The second experiment uses the same setup as the first, with the only change being that the single pillar obstacle is replaced by a wide wall with two narrow slits (Figure~\ref{fig:cloth_double_slit}). This adds a narrow passage problem and also demonstrates the utility of the progress detection filter. In this example the local controller is trying to move the deformable object straight forward, but with the wall in the way it is unable to make progress; the local controller cannot explicitly go around obstacles. This experiment shows comparable planning time, but it takes longer to smooth the resulting path (as expected given that the virtual elastic band forward propagation takes longer near obstacles). The local controller is again able to complete the task after invoking the planer a single time on all 100 trials.

\begin{table}[t]
\centering
\caption{Deadlock prediction parameters}
\label{tab:deadlock_param_table}
\begin{tabular}{lcc}
\noalign{\smallskip}\hline\noalign{\smallskip}
Prediction Horizon                  & $\predictionhorizon$          &    10 \\
Band Annealing Factor               & $\bandlengthannealing$        &   0.3 \\
History Window                      & $\historywindow$              &   100 \\
Error Improvement Threshold         & $\errorprogressthreshold$     &     1 \\
Configuration Distance Threshold    & $\motionprogressthreshold$    &  0.03 \\
\noalign{\smallskip}\hline
\end{tabular}
\end{table}

\begin{table}[t]
\centering
\caption{Distance and planner parameters}
\label{tab:rrt_param_table}
\begin{tabular}{lcc}
\noalign{\smallskip}\hline\noalign{\smallskip}
Goal Bias                           & $\goalbias$                   &     0.1 \\
Workspace Goal Radius               & $\goalreachradius$            &     0.02 \\
Best Nearest Radius                 & $\bestneardist$               &     0.001 \\
Band Distance Scaling Factor        & $\banddistscale$              & $10^{-6}$ \\
Maximum Band Points                 & $\maxbandpoints$              & 500 \\
\noalign{\smallskip}\hline
\end{tabular}
\end{table}

\subsection{Moving a Rope Through a Maze}

\begin{figure*}[t]
    \centering
    \includegraphics[width=\textwidth]{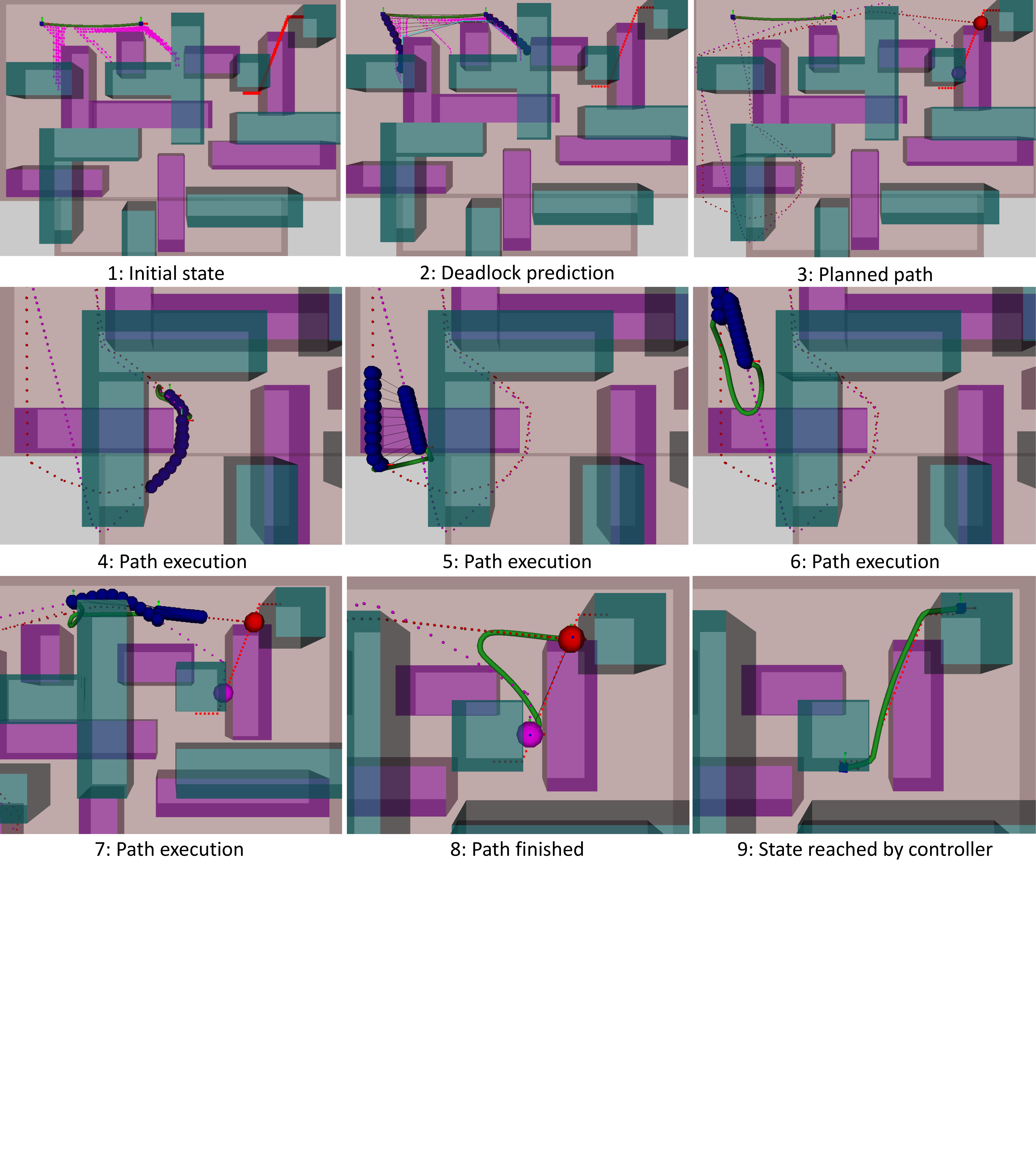}
    \vspace{-2in}
    \caption{Sequence of snapshots showing the execution of the third experiment. The rope is shown in green starting in the top left corner, the grippers are shown in blue, and the target points are shown in red in the top right corner. The maze consists of top and bottom layers (green and purple, respectively). The rope starts in the bottom layer and must move to the target on the top layer through an opening (bottom left or bottom right).}
    \label{fig:rope_maze}
\end{figure*}

In the third task, the robot must navigate a rope through a three-dimensional maze before aligning the rope with a line traced on the floor (Figure~\ref{fig:rope_maze}). This scenario is meant to represent tasks such as moving a heavy cable through a construction zone without crane access. In this task, the correspondences between the target points $\deformtarget$ and the deformable object points $\deformconfig$ are fixed in advance, thus the CalculateCorrespondences() function does not have to do any work, as shown in Table~\ref{tab:control_statistics}. Task error $\errorfunction$ is defined in the same way as in the first two experiments. Again the planner is invoked a single time per trial, but planning and smoothing times are longer than the previous tasks. This is a function of the size of the environment rather than any particular difference in the difficulty of performing the planning or smoothing. The planner finds a feasible path in 4.2s on average, suggesting that our method can maintain fast planning times, even in larger environments with many more obstacles.

\begin{table*}[t]
\centering
\caption{Planning statistics for the first plan for each example task in simulation, averaged across 100 trials. Standard deviation is shown in brackets.}
\label{tab:planning_statistics}
\begin{tabular}{lccccc|cccc}
\hline
    & \multicolumn{5}{c}{\textbf{RRT Planning}} & \multicolumn{4}{|c}{\textbf{Smoothing}} \\
    & Samples & States & \parbox{0.3in}{\centering NN\\Time (s)} & \parbox{0.55in}{\centering Validity\\Checking\\Time (s)} & \parbox{0.3in}{\centering Total\\Time (s)} & Iterations & \parbox{0.55in}{\centering Validity\\Checking\\Time (s)} & \parbox{0.75in}{\centering Visibility\\Deformation\\Time (s)}& \parbox{0.3in}{\centering \smallskip Total\\Time (s) \smallskip} \\
\hline
Single Pillar                       & \parbox{0.37in}{\centering{ 158 \\ {[121]}}} & \parbox{0.37in}{\centering{1182 \\ {[804]}}} & \parbox{0.4in}{\centering{$\sim$0.0 \\{[$\sim$0.0]}}} & \parbox{0.4in}{\centering{0.6 \\{[0.5]}}} & \parbox{0.3in}{\centering{0.6 \\{[0.5]}}} &  500 & \parbox{0.5in}{\centering{0.8 \\{[1.2]}}} & \parbox{0.4in}{\centering{       1.6 \\{[0.2]}}}       & \parbox{0.3in}{\smallskip\centering{2.4 \\{[1.2]}}\smallskip} \\

Double Slit                         & \parbox{0.37in}{\centering{ 478 \\ {[353]}}} & \parbox{0.37in}{\centering{2124 \\{[1428]}}} & \parbox{0.4in}{\centering{$\sim$0.0 \\{[$\sim$0.0]}}} & \parbox{0.4in}{\centering{0.7 \\{[0.8]}}} & \parbox{0.3in}{\centering{0.7 \\{[0.8]}}} &  500 & \parbox{0.5in}{\centering{2.5 \\{[2.6]}}} & \parbox{0.4in}{\centering{$\sim$ 0.0 \\{[$\sim$0.0]}}} & \parbox{0.3in}{\smallskip\centering{2.5 \\{[2.6]}}\smallskip} \\

Rope Maze                           & \parbox{0.37in}{\centering{4796 \\{[1613]}}} & \parbox{0.37in}{\centering{9926 \\{[3760]}}} & \parbox{0.4in}{\centering{      0.1 \\{[$\sim$0.0]}}} & \parbox{0.4in}{\centering{4.0 \\{[1.7]}}} & \parbox{0.3in}{\centering{4.2 \\{[1.8]}}} & 1500 & \parbox{0.5in}{\centering{6.4 \\{[3.9]}}} & \parbox{0.4in}{\centering{$\sim$ 0.0 \\{[$\sim$0.0]}}} & \parbox{0.3in}{\smallskip\centering{6.5 \\{[3.9]}}\smallskip} \\

\parbox{0.6in}{Repeated\\Planning} & \parbox{0.37in}{\centering{ 54  \\  {[46]}}}  & \parbox{0.37in}{\centering{153  \\ {[147]}}} & \parbox{0.4in}{\centering{$\sim$0.0 \\{[$\sim$0.0]}}} & \parbox{0.4in}{\centering{0.1 \\{[0.1]}}} & \parbox{0.3in}{\centering{0.1 \\{[0.1]}}} &  500 & \parbox{0.5in}{\centering{1.4 \\{[0.9]}}} & \parbox{0.4in}{\centering{$\sim$ 0.0 \\{[$\sim$0.0]}}} & \parbox{0.3in}{\smallskip\centering{1.4 \\{[0.9]}}\smallskip} \\
\hline
\end{tabular}
\end{table*}

\begin{figure*}[t]
    \centering
    \includegraphics[width=\textwidth]{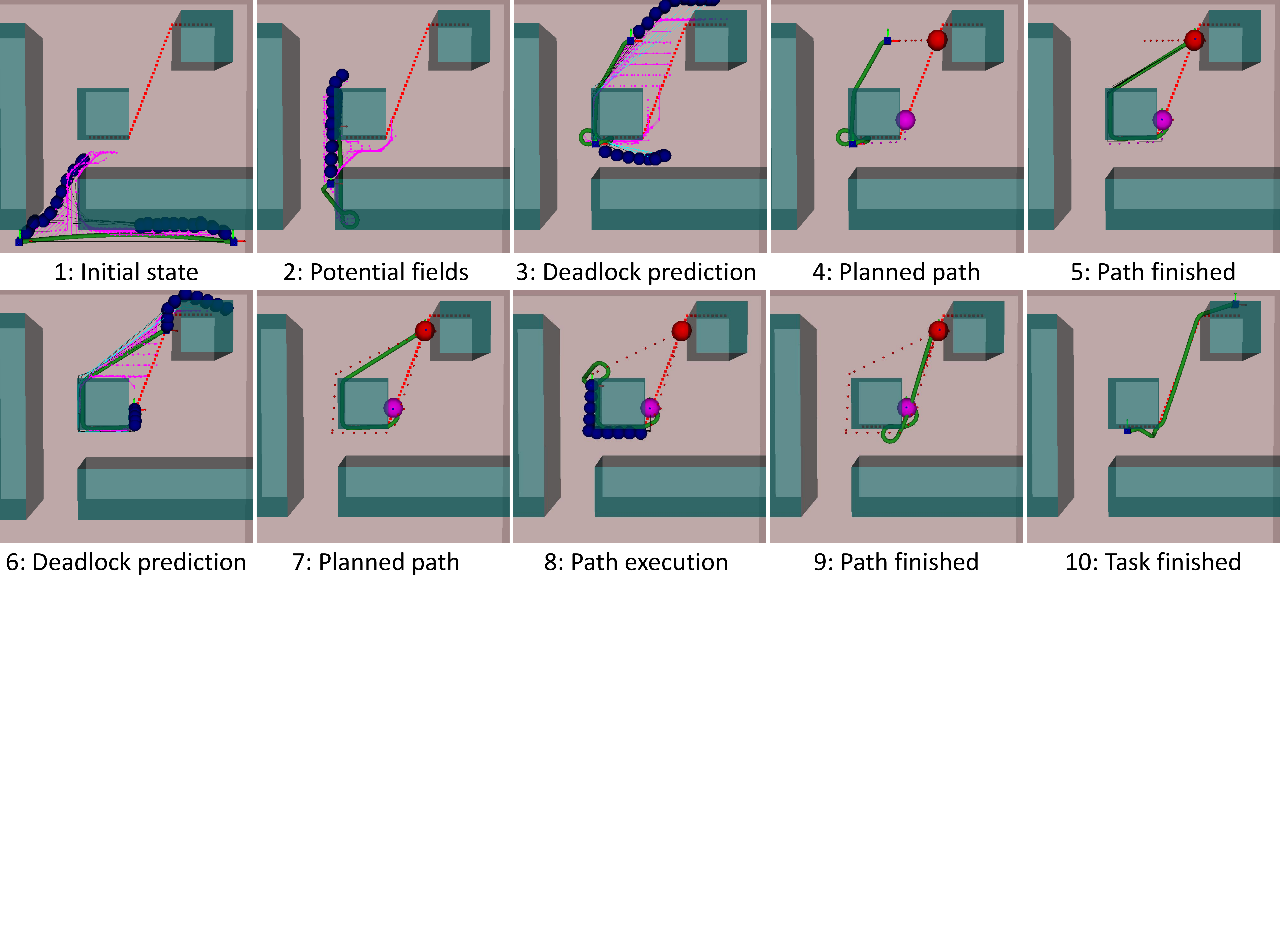}
    \vspace{-2in}
    \caption{Sequence of snapshots for the fourth experiment. We use the same colors as the previous experiment (Figure~\ref{fig:rope_maze}), but in this example the local controller gets stuck twice, in panels 3 and 6. In panel 7 the global planner finds a new neighbourhood that is distinct from previously-tried neighbourhoods.}
    \label{fig:repeated_planning}
\end{figure*}

\subsection{Repeated Planning}

The fourth task is a variant of the third, with the start configuration of the rope moved near the goal region on the top layer of the maze and a longer rope. This task has the most potential for a planned path to move the deformable object into a configuration from which the local controller cannot finish the task by wrapping the rope around an obstacle near the goal. For this experiment we reduce the size of the planning arena to only the goal area, and the immediate surroundings on the top layer (Figure~\ref{fig:repeated_planning}). From this starting position, the planner is more likely to find the incorrect neighborhood for the local controller, which corresponds to placing the rope into the wrong homotopy class, on the first attempt. We emphasize that the correct homotopy class is unknown, as we assume no information is given about the connectivity of the target points. Thus our method must discover the correct homotopy class by trail-and-error, invoking the planner when the deadlock prediction determines the controller will be stuck.

In 71 of the 100 trials, the planner was invoked twice, in 13 other trials it was invoked three times, and in 2 trials it was invoked four times. These additional planning and smoothing stages took on average an additional 6.6 seconds, but the task was completed successfully in all 100 trials. This experiment suggests that our framework is able to effectively explore different band neighborhoods until the correct one is found, enabling the local controller to finish the task, even when the initial configuration is adversarial.

\subsection{Computation Time}


To verify the practicality of our deadlock prediction algorithm and virtual elastic band approximation, we gathered data comparing computation time for these components to the local controller by itself, and to using the Bullet simulator. Table~\ref{tab:control_statistics} shows the average times per iteration for the local controller and deadlock prediction algorithms, averaged across all trials of all experiments. As expected, adding in the deadlock prediction step does increase computation time, but the overall control loop is still fast enough for practical use.

\begin{table}[t]
\centering
\caption{Local controller and deadlock prediction avg. computation time per iteration for each type of deformable object, averaged across all trials.}
\label{tab:control_statistics}
\begin{tabular}{lccc}
\noalign{\smallskip}\hline\noalign{\smallskip}
& 
\parbox{1.04in}{\centering Calculate\\Correspondences()\\Time (s)} &
\parbox{0.62in}{\centering Predict\\Deadlock() Time (s)} &
\parbox{0.64in}{\centering Local\\Controller Time (s)} \\
\noalign{\smallskip}\hline\noalign{\smallskip}
Cloth   & 0.0114 & 0.0077 & 0.0126 \\
Rope    & 0      & 0.0119 & 0.0023 \\
\noalign{\smallskip}\hline
\end{tabular}
\end{table}

\begin{table}[t]
\centering
\caption{Average computation time to compute the effect of a gripper motion.}
\label{tab:prediction_statistics}
\begin{tabular}{lcc}
\noalign{\smallskip}\hline\noalign{\smallskip}
& 
\parbox{0.7in}{\centering Bullet\\Simulation\\Time (ms)} &
\parbox{1.1in}{\centering Virtual Elastic\\Band Propagation\\Time (ms)} \\
\noalign{\smallskip}\hline\noalign{\smallskip}
Cloth   & 36.12 & 0.19 \\
Rope    &  3.19 & 0.58 \\
\noalign{\smallskip}\hline
\end{tabular}
\end{table}


Table~\ref{tab:prediction_statistics} shows a comparison between the average time needed to compute the virtual elastic band propagation for a gripper motion and the time needed to reliably simulate a gripper motion with the Bullet simulator. Note that the amount of time required for the simulator to converge to a stable estimate depends on many conditions, including what object is being simulated. Through experimentation we determined that 4 simulation steps were adequate for rope and 10 for cloth. Comparing the time needed to do this simulation to the time needed to forward propagate a virtual elastic band, we see that our approximation is indeed faster by an order of magnitude for rope, and by two orders of magnitude for cloth. This result reinforces the importance of using a simplified model, such as the virtual elastic band, within the planner---this model, while not as accurate as a simulation, allows us to evaluate motions much faster.


\begin{figure*}[t]
    \centering
    \includegraphics[width=\textwidth]{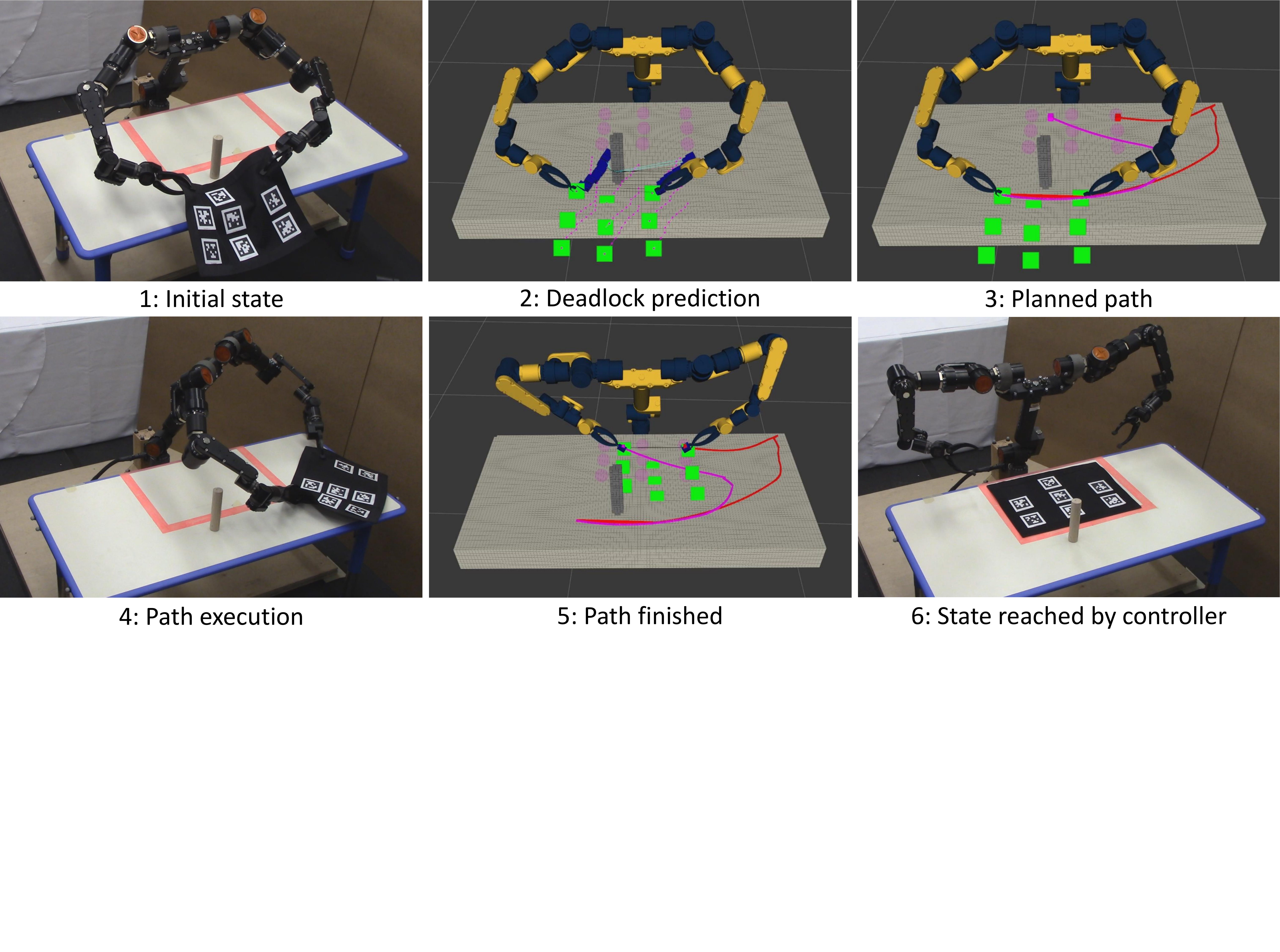}
    \vspace{-1.7in}
    \caption{Cloth placemat task. The placemat starts on the far side of an obstacle and must be aligned with the pink rectangle near the robot.}
    \label{fig:cloth_placemat}
\end{figure*}

\section{Physical Robot Experiment and Results}
\label{sec:live_robot}

In order to show that our method is practical for a physical robotic system, not only free floating end-effectors, we set up a task similar to the single pillar task (Sec.~\ref{sec:single_pillar}) with a dual-arm robot. It also shows that while our methods strong assumptions about the ability to perceive the deformable object in Sec.~\ref{sec:main_problem_statement} (in particular no occlusions and no sensor noise), our framework is still able to perform meaningful tasks when those assumptions are violated. In this task the robot must align a cloth placemat inside of the pink rectangle, going around an obstacle in the process (Fig.~\ref{fig:cloth_placemat}).

\subsection{Experiment Setup}

\subsubsection{Robotic Platform:}
Val is a stationary robotic platform with a 2-DOF torso, two 7-DOF arms, and a rotary pincer per arm. As in the simulated environments it is assumed that Val is already holding the cloth, leaving 16 DOF to be controlled and planned for ($\cspace_r = \reals^{16})$.

\subsubsection{Cloth Perception:}
\label{sec:cloth_perception}

The placemat is $0.33\text{m} \times 0.46\text{m}$ which we discretize into a $3 \times 3$ grid. As tracking of deformable objects is a difficult problem, and out of scope of this paper, we instead use fiducials to track the configuration of the cloth. Two of the points are tracked using the position of the grippers; the other 7 points are tracked with AprilTags (\cite{olson2011tags}) and a Kinect V2 RGB-D sensor (\cite{iai_kinect2}).

\begin{figure}[t]
    \centering
    \includegraphics[width=0.9\columnwidth]{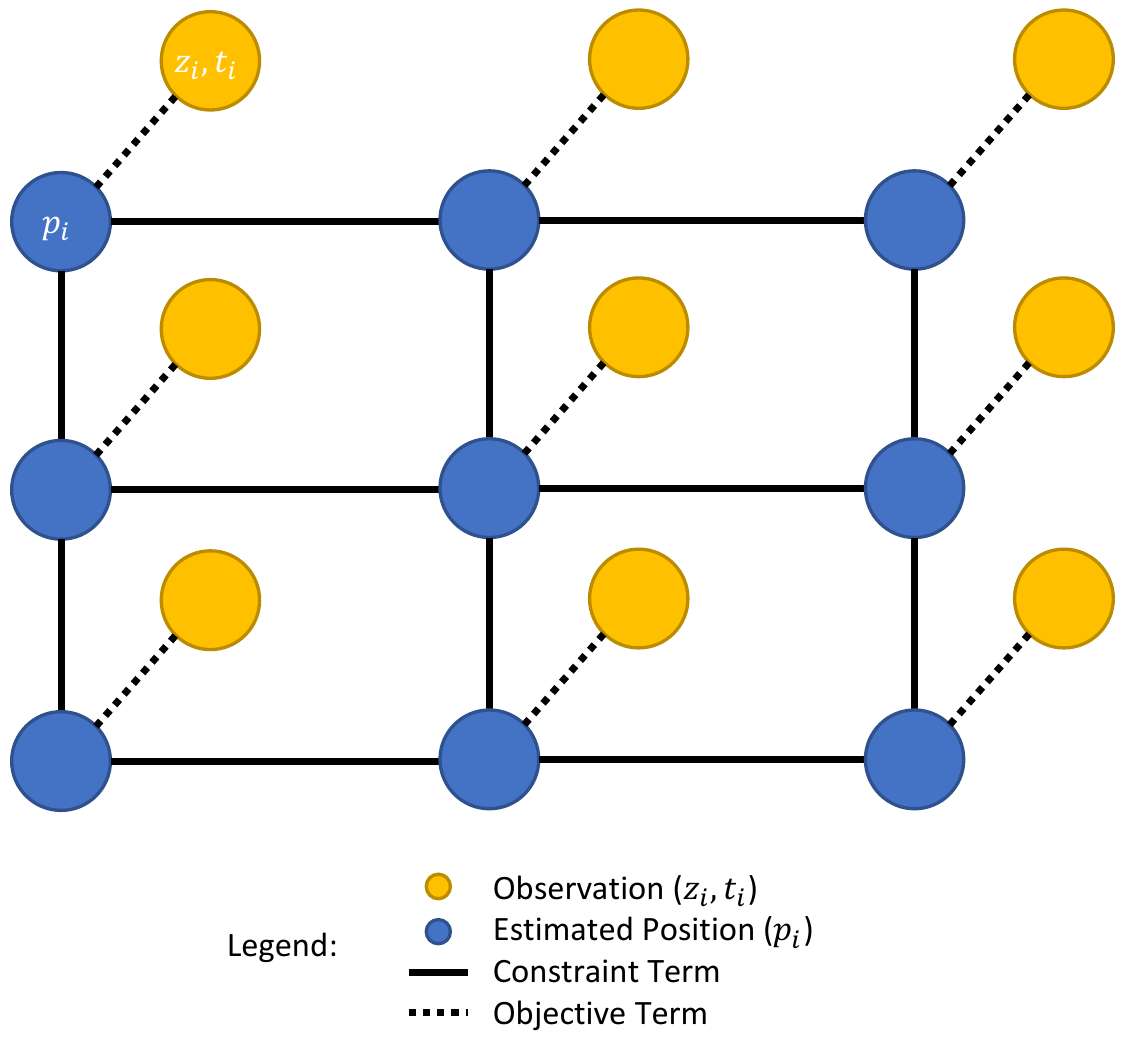}
    \caption{Constraint and objective graph for Eq.~\eqref{eqn:cloth_estimation}. Note that not all constraints are shown to avoid clutter; every estimated position has a constraint between itself and every other estimated position.}
    \label{fig:cloth_estimation}
\end{figure}

In order to address occlusions and noisy data, we filter the raw observations using a set of objective terms, and a set of constraints (see Fig.~\ref{fig:cloth_estimation}). Denote $z_i$ as the last observed position of point $i$, and denote $t_i$ as the last time point $i$ was observed. Then we add objective terms to pull the cloth estimate towards the observations, combined with constraints between each pair of points to ensure that the estimate is plausible:
\begin{equation}
\begin{aligned}
    \deformconfig(t) = &\argmin_{\{ p_i \}} 
            & & \sum_i e^{-K_T(t - t_i)} \| p_i - z_i \|^2 \\
            & \text{subject to}
            & & \| p_i - p_j \|^2 \leq K_L d_{ij}^2 \enspace \forall i, j \text{ s.t. } i \neq j \enspace .
\end{aligned}
\label{eqn:cloth_estimation}
\end{equation}
$K_T$ and $K_L$ are task defined scale factors which we set to $1.5$ and $1.0001$ respectively for this task.

\subsection{Experiment Results}

We use the same deadlock, distance, and planner parameters as used in the simulation experiments, performing 500 smoothing iterations once a path is found. We constrain the rotation of the end-effectors to stay within 1.6 radians of their starting orientation during the planning process as well as constrain the grippers to stay close to the table. This forces the planner to move the placemat around the obstacle rather than over the obstacle. Last, we also introduce planning restarts (\cite{Wedge2008}) into the planning process in order to address the greater complexity added by using a 16-DOF robot and the relatively strict workspace constraints; the restart timeout we set is 60 seconds.

\begin{table*}[t]
\centering
\caption{Planning statistics for the cloth placemat example, averaged across 100 trials. Standard deviation is shown in brackets.}
\label{tab:live_robot_stats}
\begin{tabular}{cccccc|cccc}
\hline
\multicolumn{6}{c}{\textbf{RRT Planning}} & \multicolumn{4}{|c}{\textbf{Smoothing}} \\
Samples & 
States & 
\parbox{0.3in}{\centering NN\\Time (s)} & 
\parbox{0.55in}{\centering Validity\\Checking\\Time (s)} & 
\parbox{0.5in}{\centering Random\\Restarts} &
\parbox{0.3in}{\centering Total\\Time (s)} & 
Iterations & 
\parbox{0.55in}{\centering Validity\\Checking\\Time (s)} & 
\parbox{0.75in}{\centering Visibility\\Deformation\\Time (s)}& 
\parbox{0.3in}{\centering \smallskip Total\\Time (s) \smallskip} \\
\hline
\parbox{0.45in}{\centering 83041\\{[83677]}} &
\parbox{0.4in}{\centering 8438\\{[6182]}} &
\parbox{0.3in}{\centering 4.5\\{[4.9]}} &
\parbox{0.4in}{\centering 44.1\\{[44.5]}} &
\parbox{0.3in}{\centering 0.5\\{[0.9]}} &
\parbox{0.4in}{\centering 50.0\\{[50.9]}} &
500 &
\parbox{0.3in}{\centering 3.6\\{[1.1]}} &
\parbox{0.4in}{\centering 0.1\\{[$\sim$0.0]}} &
\parbox{0.3in}{\smallskip\centering 3.6\\{[1.1]} \smallskip} \\
\hline
\end{tabular}
\end{table*}

\begin{figure}[t]
    \centering
    \includegraphics[width=\columnwidth]{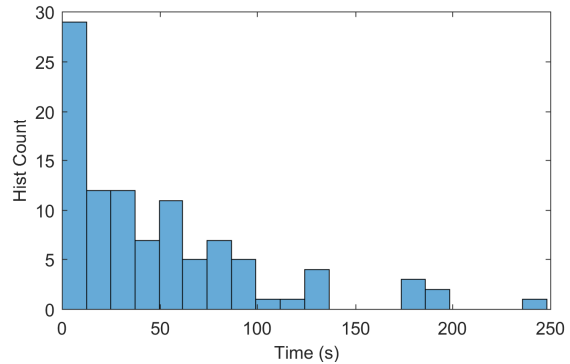}
    \caption{Histogram of planning times across 100 trials for the cloth placemat experiment.}
    \label{fig:placemat_planning_time}
\end{figure}

Table~\ref{tab:live_robot_stats} shows the planning statistics across 100 planning trials with identical starting configurations, but different random seeds. On average planning and smoothing takes less than 60 seconds, with forward kinematics and collision checking dominating the planning time. The restart timeout was unused in 68 out of 100 trials, with the other 32 trials requiring a total of 50 restarts between them. Fig.~\ref{fig:placemat_planning_time} shows that the planning time follows a ``heavy tail'' distribution typical of sampling-based planners.

Our overall framework is able to complete this task as shown in Fig.~\ref{fig:cloth_placemat}. As in the simulated version of this task, we are able to predict deadlock before the robot gets stuck, plan and execute a path to a new neighbourhood, and then use the local controller to finish the task.

\section{Discussion and Conclusion}
\label{sec:discussion}

We have presented a method to interleave global planning and local control for deformable object manipulation that does not rely on high-fidelity modeling or simulation of the object. Our method combines techniques from topologically-based motion planning with a sampling-based planner to generate gross motion of the deformable object. The purpose of this gross motion is not to achieve the task alone, but rather to move the object into a position from which the local controller is able to complete the task. This division of labor enables each component to focus on their strengths rather than attempt to solve the entire problem directly. We also presented a probabilistic completeness proof for our planner which does not rely on either a steering function or choosing controls at random, and addresses our underactuated system. As part of our framework, we introduced a novel deadlock prediction algorithm to determine when to use the local controller and when to use the global planner.

Our experiments demonstrate that our framework is able to be applied to several interesting tasks for rope and cloth, including an adversarial case where we set up the planner to fail on the first attempt. For the simulated tasks, our framework is able to succeed at each task 100/100 times, with average planning and smoothing time under 4 seconds for 3 tasks, and under 11 seconds for the larger environment. The physical robot experiment shows that our framework can be used for practical tasks in the real world, with planning and smoothing taking less than 60 seconds on average. This experiment also shows that our methods can function despite noisy and occluded perception of the deformable object.

\subsection{Parameter Selection}
There are several parameters in both the local controller and the global planner that can have a large impact on the performance of our method. In particular, if the local controller is prone to oscillations \rev{(Sec.~\ref{sec:ctrl_params})}, this can cause the deadlock prediction algorithm to incorrectly predict that the local controller will get stuck, leading to an unnecessary planning phase. In the worse case, this can cause the global planner to be unable to find an acceptable path due to the blacklisting procedure. One interesting direction of future research is how to perform reachability analysis for deformable objects in general, in particular when a high-fidelity model of the deformable object is not available. \rev{In practice we found that increasing the prediction horizon $\predictionhorizon$ and prediction annealing factor $\bandlengthannealing$ was not useful as the prediction accuracy degrades quickly. We did have to tune the history window $\historywindow$ and thresholds $\errorprogressthreshold$, $\motionprogressthreshold$ against each other. Error improvement threshold $\errorprogressthreshold$ needs to be set relative to the definition of task error $\errorfunction$, while $\motionprogressthreshold$ is more sensitive to oscillations. If $\motionprogressthreshold$ is too small, then the system will fail to detect that the controller is stuck in a poor local minima. If these thresholds are too high or $\historywindow$ is too low, then false positives were common near the end of the table coverage tasks.}

\rev{For the global planner, we found that the goal bias $\goalbias$ has a similar effect on planning time as a standard RRT; values in the range $[0.05, 0.15]$ produced similar planning times for our experiments}. In addition, if $\banddistscale$ is not small, then nearest neighbour checks can become very expensive. In practice distances in band space are used to disambiguate between nodes that are at nearly identical configurations in robot configuration space. This happens when multiple nodes connect to the position goal $\eepositiongoal$, but their bands are similar to a blacklisted band. One potential way to make distances in band space more informative would be to develop a way to sample interesting band configurations.

\subsection{Limitations}
We made a choice to favor speed over model accuracy. As a consequence, there are several issues that our method does not address. In particular environments with ``hooks'' can cause problems due to our approximation methods; the virtual elastic band we use for constraint checking and planning assumes \rev{(1)} that there is no minimum length of the deformable object \rev{and (2) there are no holes in the deformable object. These assumptions mean} that our planner cannot detect cases where the \rev{slack material or a hole} can get snagged on corners or hooks, preventing the motion plan from being executed. One way this can be mitigated is by using a more accurate model (at the cost of speed and task-specific tuning). Other potential solutions include online modeling methods such as~\cite{Hu2018deformable_gpr}, or learning which features of the workspace can lead to highly inaccurate approximations and planning paths that avoid those areas. In addition we have no explicit method to avoid twisting or knot-tying behavior. While shortcut smoothing can potentially mitigate the worst effects, avoiding such cases is not something that is within the scope of this work. \rev{Similarly, we don't have any explicit consideration for achieving a task that requires knot-tying or twisting; while some other local controller may be able to perform these tasks from a suitable starting state, we have not investigated this option.} Last, we cannot guarantee that we can achieve any given task in general; while our blacklisting method is designed to encourage exploration of the state space, it also has the potential to block regions of the state space from which the local controller can achieve the task. \rev{Defining a set of tasks which our framework can successfully perform is not practical given the limited set of assumptions we are making about the deformable object.} Despite these limitations we find that our framework is able to reliably perform complex tasks where neither planning nor control alone are sufficient. In future work we plan to address these weaknesses, in particular the snagging and twisting limitations which are artifacts of our approximation methods. We also seek to extend our framework to a broader range of tasks, beyond coverage and point matching applications.






\bibliographystyle{SageH}
\bibliography{references_strings_abbrev,references_dale}

\appendix
\section*{Appendices}

\section{Local Controller}
\label{apx:local_control}

This appendix provides the details of each component of the local controller (Alg.~\ref{alg:local_controller}). The three main sections determine: (1) Which direction to manipulate the deformable object in order to reduce task error; (2) Adjustments to help avoid overstretch of the deformable object; and (3) Determining the best direction to move the robot to achieve (1) and (2). We discuss each section in turn.

\subsection{Reducing error}

Detrmining which direction to manipulate the deformable object in order to reduce task error is done in three steps (Algorithms \ref{alg:calculate_correspondences} and \ref{alg:follow_nav_function}).

Each task defines a navigation function for every target point $\deformtarget_i$ using Dijkstra's algorithm. In general there is not a one to one mapping between $\deformtarget$ and $\deformconfig$; at every timestep, for every target point $\deformtarget_i$, we recalculate which point on the deformable object $\deformconfig_j$ is closest, using the results from Dijkstra's algorithm to measure distance (Alg.~\ref{alg:calculate_correspondences}). These individual results are then aggregated in Algorithms~\ref{alg:follow_nav_function} to define the best direction to manipulate the deformable object in order to reduce error, and the relative importance of doing so for each point on the deformable object. The directions each navigation function indicates are added together to define the overall direction to manipulate a point (Alg.~\ref{alg:follow_nav_function} line 5). For the importance factors $\pseudoinverseweight_j$, we take only the largest distance that $\deformconfig_j$ would have to move as a way to mitigate discretization effects (Alg.~\ref{alg:follow_nav_function} line 6).

\FloatBarrier

\begin{algorithm}[t]
\caption{CalculateCorrespondences$(\deformconfig, \deformtarget)$}
\begin{algorithmic}[1]
    \State $\correspondences = [ \emptyset ]_{1 \times \numtargetpoints}$
    \For {$i \in \{ 1, 2, \dots, \numtargetpoints \}$}
        \State $j \gets \argmin_{j \in \{ 1, 2, \dots, \numdeformpoints \}} d_\textrm{Dijkstras}(\deformtarget_i, \deformconfig_j)$
        \State $d \gets d_\textrm{Dijkstras}(\deformtarget_i, \deformconfig_j)$
        \State $\correspondences[j] \gets \{ \correspondences[j] \cup (j, d)\}$
    \EndFor
    \State \Return $\correspondences$
\end{algorithmic}
\label{alg:calculate_correspondences}
\end{algorithm}

\begin{algorithm}[t]
\caption{FollowNavigationFunction$(\deformconfig, \correspondences)$}
\begin{algorithmic}[1]
    \State $\deformvelocity_e \gets \boldsymbol 0_{3\numdeformpoints \times 1}$
    \State $\pseudoinverseweight_e \gets \boldsymbol 0_{\numdeformpoints \times 1}$
    \For {$j \in \{ 1, 2, \dots, \numdeformpoints \}$}
        \For {$(j, d) \in \correspondences[j]$}
            \State $\deformvelocity_{e,i} \gets \deformvelocity_{e,i} +$ DijkstrasNextStep$(\deformconfig_i, j)$
            \State $\pseudoinverseweight_{e,j} \gets \max(\pseudoinverseweight_{e,j}, d)$
        \EndFor
    \EndFor
    \State \Return $\deformvelocity_e, \pseudoinverseweight_e$
\end{algorithmic}
\label{alg:follow_nav_function}
\end{algorithm}

\subsection{Stretching correction}

\begin{algorithm}[t]
\caption{StretchingCorrection$(\relaxeddistancematrix, \maxstretchfactor, \deformconfig)$ \\
         (adapted from \cite{McConachie2018}}
\begin{algorithmic}[1]
    \State $E \gets$ EuclidianDistanceMatrix$(\deformconfig)$
    \State $\deformvelocity_s \gets \boldsymbol 0_{3\numdeformpoints \times 1}$, $\pseudoinverseweight_s \gets \boldsymbol 0_{\numdeformpoints \times 1}$
    \For{$i \in \{1,2,\dots,\numdeformpoints \}$}
        \For{$j \in \{i+1,\dots,\numdeformpoints \}$}
            \If{$E_{i,j} > \maxstretchfactor D_{i,j}$}
                \State $\Delta_{i,j} \gets E_{i,j} - D_{i,j}$
                \State $v \gets \Delta_{i,j}(\deformconfig_j - \deformconfig_i)$
                \State $\deformvelocity_{s,i} \gets \deformvelocity_{s,i} + \frac{1}{2}v$
                \State $\deformvelocity_{s,j} \gets \deformvelocity_{s,j} - \frac{1}{2}v$
                \State $\pseudoinverseweight_{s,i} \gets \max (\pseudoinverseweight_{s,i}, \Delta_{i,j})$
                \State $\pseudoinverseweight_{s,j} \gets \max (\pseudoinverseweight_{s,j}, \Delta_{i,j})$
            \EndIf
        \EndFor
    \EndFor
    \State \Return $\deformvelocity_s, \pseudoinverseweight_s$
\end{algorithmic}
\label{alg:stretching_correction}
\end{algorithm}

\begin{algorithm}[t]
\caption{CombineTerms$(\deformvelocity_e, \pseudoinverseweight_e, \deformvelocity_s, \pseudoinverseweight_s, \stretchingcorrectionweightfactor)$ \\
         (adapted from \cite{McConachie2018}}
\begin{algorithmic}[1]
    \For{$i \in \{ 1,2,\dots,\numdeformpoints \}$}
        \State $\deformvelocity_{d,i} \gets \deformvelocity_{s,i} + \left( \deformvelocity_{e,i} - \proj_{\deformvelocity_{s,i}} \deformvelocity_{e,i} \right)$
        \State $\pseudoinverseweight_{d,i} \gets \stretchingcorrectionweightfactor \pseudoinverseweight_{s,i} + \pseudoinverseweight_{e,i}$
    \EndFor
    \State \Return $\deformvelocity_d, \pseudoinverseweight_d$
\end{algorithmic}
\label{alg:combine_terms}
\end{algorithm}

Our algorithm for stretching correction is similar to that found in~\cite{Berenson2013}, with the addition of a weighting term $\stretchingcorrectionweightfactor$, and a change in how we combine error correction and stretching correction. We use the StretchingCorrection() function (Alg.~\ref{alg:stretching_correction}) to compute $\deformvelocity_s$ and $\pseudoinverseweight_s$ based on a task-defined stretching threshold $\pseudoinverseweight_s \geq 0$. First we compute the distance between every two points on the object and store the result in $E$. We then compare $E$ to $D$ which contains the relaxed lengths between every pair of points. If any two points are stretched by more than a factor of $\pseudoinverseweight_s$, we attempt to move the points closer to each other. We use the same strategy for setting the importance of this stretching correction $\pseudoinverseweight_s$ as we use for error correction. When combining stretching correction and error correction terms (Alg.~\ref{alg:combine_terms}) we prioritize stretching correction, accepting only the portion of the error correction that is orthogonal to the stretching correction term for each point. $\stretchingcorrectionweightfactor$ is used to define the relative scale of the importance factors $\pseudoinverseweight_e$ and $\pseudoinverseweight_s$

\FloatBarrier

\begin{table*}[t]
\centering
\caption{Controller parameters}
\label{tab:controller_param_table}
\begin{tabular}{lcccc}
\hline\noalign{\smallskip}
                                        &                                   & \parbox{0.5in}{\centering Simulated\\Cloth\\Trials} 
                                                                            & \parbox{0.5in}{\centering Simulated\\Rope\\Trials}
                                                                            & \parbox{0.5in}{\centering Physical\\Robot} \\
\noalign{\smallskip}\hline\noalign{\smallskip}
Servoing max gripper velocity           & $\maxgrippervelservo$             &   0.2 &   0.2 &    0.3 \\
Obstacle avoidance max gripper velocity & $\maxgrippervelobstacle$          &   0.2 &   0.2 &      - \\
Max robot velocity                      & $\dot \config_{r,\textrm{max}}$   &     - &     - &    1.5 \\
Obstacle avoidance scale factor         & $\beta$                           &   200 &  1000 &      - \\
Max stretching factor                   & $\maxstretchfactor$               &  1.15 &  1.17 &   1.01 \\
Stretching correction weight factor   & $\stretchingcorrectionweightfactor$ &  2000 &  2000 &   2000 \\
Obstacle avoidance buffer               & $d_\textrm{buffer}              $ &     - &     - &   0.08 \\
\rev{Workspace discretization (m)}      &                                   &  \rev{0.02} &  \rev{0.05} &   \rev{0.02} \\
\hline
\end{tabular}
\end{table*}

\subsection{Finding the best robot motion}

Given a desired deformable object velocity $\deformvelocity_d$ and relative importance weights $\pseudoinverseweight_d$, we want to find the robot motion that best achieves $(\deformvelocity_d, \pseudoinverseweight_d$. I.e.
\begin{equation}
\begin{aligned}
    & \argmin_{\robotvelocity } 
        & & \| f(\robotconfig, \deformconfig, \robotvelocity) - \deformvelocity_d \|_{\pseudoinverseweight_d} \\
    &\text{subject to}
        & & \| \robotvelocity \| \leq \maxrobotvel \\
    &   & & \left(\robotconfig + \robotvelocity\right) \in \cfree_r \enspace .
\end{aligned}
\label{eqn:controller_minimization_problem_apx}
\end{equation}
In general, $f(\dots)$ is not known. For our controllers we use a Jacobian based approximation
\begin{equation}
    f(\robotconfig, \deformconfig, \robotvelocity) \approx J_d \robotvelocity
\end{equation}
from \cite{McConachie2018} \textsection{V-C}. 

Our method for ensuring the robot stays in $\cfree_r$ is different, depending on which robot we are using.

\subsubsection{Simulated experiments:}

For the simulated experiments, we first solve Eq.~\eqref{eqn:controller_minimization_problem_apx} using our Jacobian approximation:
\begin{equation}
\begin{aligned}
    \deformablemodelbackwardfunction_{\tanse3}(\deformvelocity, \pseudoinverseweight) =
                &\argmin_{\robotvelocity }  & & \| J_d \robotvelocity - \deformvelocity \|^2_{\pseudoinverseweight} \\
                &\text{subject to}          & & \| \robotvelocity \|^2 \leq \maxgrippervelservo^2
    \label{eqn:jacobianbackwardfunction_sim}
\end{aligned}
\end{equation}
where $\maxgrippervelservo$ is the maximum velocity for each individual end-effector (Alg.~\ref{alg:find_best_robot_motion_simulation}).

In order to guarantee that the grippers do not collide with any obstacles, we use the same strategy from~\cite{Berenson2013}, smoothly switching between collision avoidance and other objectives (see Alg.~\ref{alg:obstaclerepulsion}). For every gripper $\gripperindex$ and an obstacle set $\obstacle$ we find the distance $d_\gripperindex$ to the nearest obstacle, a unit vector $\dot x_{p_\gripperindex}$ pointing from the obstacle to the nearest point on the gripper, and a Jacobian $J_{p^\gripperindex}$ between the gripper's DoF and the point on the gripper as shown in Alg.~\ref{alg:proximity}. We then project the servoing motion from Eq.~\eqref{eqn:jacobianbackwardfunction_sim} into the null space of the avoidance motion using the null space projector $\left(\eye - J_{p^g}^+ J_{p^g} \right)$. $\beta > 0$ sets the rate at which we change between servoing and collision avoidance objectives. $\maxgrippervelobstacle > 0$ is an internal parameter that sets how quickly we move the robot away from obstacles.

\begin{algorithm}[t]
\caption{FindBestRobotMotionSim$(\robotconfig, \deformconfig, \deformvelocity_d, \pseudoinverseweight_d)$}
\label{alg:find_best_robot_motion_simulation}
\begin{algorithmic}[1]
    \State $\robotcommandvel \gets \deformablemodelbackwardfunction_{\tanse3} (\deformvelocity_, \pseudoinverseweight_d)$ \hfill Eq.~\eqref{eqn:jacobianbackwardfunction_sim}
    \State $\robotcommandvel \gets $ ObstacleRepulsion$(\robotcommandvel, \obstacle, \beta)$
    \State \Return $\robotcommandvel$
\end{algorithmic}
\end{algorithm}

\begin{algorithm}[t]
\caption{ObstacleRepulsion$(\obstacle, \beta)$ \\ (adapted from~\cite{McConachie2018})}
\begin{algorithmic}[1]
    \For{$\gripperindex \in \{1,2\}$}
        \State $J_{p^\gripperindex}, \dot x _{p^\gripperindex}, d_\gripperindex \gets$ Proximity$(\obstacle, \gripperindex)$
        \State $\gamma \gets e^{-\beta d_g}$
        \State $\dot \config_{r,\gripperindex,c} \gets J_{p^g}^+ \dot x_{p^g}$
        \State $\dot \config_{r,\gripperindex,c} \gets \frac{\maxgrippervelobstacle}{\| \dot \config_{r,\gripperindex,c} \|} \dot \config_{r,\gripperindex,c}$
        \State $\dot \config_{r,\gripperindex,c} \gets$ \parbox[t]{2in}{$\gamma \left( \dot \config_{r,\gripperindex,c} + \left( \eye - J_{p^g}^+ J_{p^g} \right) \dot \config_{r,\gripperindex,c} \right)$ \\ $+ (1-\gamma)\dot \config_{r,\gripperindex}$}
    \EndFor        
    \State \Return $\robotvelocity$
\end{algorithmic}
\label{alg:obstaclerepulsion}
\end{algorithm}

\begin{algorithm}[t]
\caption{Proximity$(\gripperindex, \obstacle)$ \\ (adapted from~\cite{McConachie2018})}
\label{alg:proximity}
\begin{algorithmic}[1]
    \State $d_\gripperindex \gets \infty$
    \For{$o \in \{1,2,\dots,|\obstacle|\}$}
        \State $p^\gripperindex, p^o \gets$ ClosestPoints$(\gripperindex, o)$
        \State $v \gets p^\gripperindex - p^o$
        \If{$\| v \| < d_\gripperindex$}
            \State $d_\gripperindex \gets \| v \|$
            \State $\dot x_{p^\gripperindex} \gets \frac{v}{\| v \|}$
            \State $J_{p^\gripperindex} \gets$ RobotPointJacobian$(\gripperindex, p^\gripperindex)$
        \EndIf
    \EndFor
    \State \Return $J_{p^\gripperindex}, \dot x_{p^\gripperindex}, d_\gripperindex$
\end{algorithmic}
\end{algorithm}

\begin{algorithm}[t]
\caption{FindBestRobotMotionPhys$(\robotconfig, \deformconfig, \deformvelocity_d, \pseudoinverseweight_d)$}
\label{alg:find_best_robot_motion_physical}
\begin{algorithmic}[1]
    \For {$\gripperindex \in \{ 1, 2, \dots, | \mathcal{C} | \}$}
        \State $J_{p^\gripperindex}, \dot x _{p^\gripperindex}, d_\gripperindex \gets$ Proximity$(\obstacle, \gripperindex)$
    \EndFor
    \State $\robotcommandvel \gets \deformablemodelbackwardfunction_{\reals^{16}} (\deformvelocity_, \pseudoinverseweight_d)$ \hfill Eq.~\eqref{eqn:jacobianbackwardfunction_live}
\end{algorithmic}
\end{algorithm}

\subsubsection{Physical experiments:}

For the physical robot, instead of handling collision avoidance in a post-processing step, we build the collision constraints directly into the optimization function (Alg.~\ref{alg:find_best_robot_motion_physical}). To do so, we define a set of points  $\mathcal{C} = \{c_1, c_2, \dots \}$ on the robot that must stay at least $d_\textrm{buffer}$ away from obstacles. In our implementation, this is the end-effectors, wrists, and elbows of each arm of the robot. We then use the same Proximity() function (Alg.~\ref{alg:proximity}) as the simulated robot to define an extra constraint that must be satisfied:
\begin{equation}
\begin{aligned}
    \deformablemodelbackwardfunction_{\reals^{16}}(\deformvelocity, \pseudoinverseweight) = 
        &\argmin_{\robotvelocity }  & & \| J_d \robotvelocity - \deformvelocity \|^2_{\pseudoinverseweight} \\
        &\text{subject to}          & & \hspace{1.2cm} \llap{$\robotconfig + \robotvelocity$} \in \cspace_r \\
        &                           & & \hspace{1.2cm} \llap{$\| \robotvelocity \|^2$}        \leq \dot \config_{r,\textrm{max}}^2 \\
        &                           & & \hspace{1.2cm} \llap{$\| J_r \robotvelocity \|^2$}    \leq \maxgrippervelservo^2 \\
        &                           & & \hspace{1.2cm} \llap{$\dot x_{p^\gripperindex}^T J_{p^\gripperindex} \robotvelocity$} \leq d_\gripperindex + d_\textrm{buffer} \enspace .\\
\label{eqn:jacobianbackwardfunction_live}
\end{aligned}
\end{equation}
In addition, we constrain the velocity of the robot both in joint configuration space 
$$ \| \robotvelocity \|^2 \leq \dot \config_{r,\textrm{max}}^2 $$
and the velocity of the end-effectors in $\se3$ 
$$ \| J_r \robotvelocity \|^2 \leq \maxgrippervelservo^2 \enspace .$$

To solve Equations \eqref{eqn:jacobianbackwardfunction_sim} and \eqref{eqn:jacobianbackwardfunction_live} we use the Gurobi optimizer (\cite{Gurobi2016}). Table~\ref{tab:controller_param_table} shows the parameters we use for each experiment.

\subsection{\rev{Parameter Selection}}
\label{sec:ctrl_params}

\rev{While this controller is able to perform multiple coverage tasks successfully, it can be prone to oscillations in three circumstances in particular. First, If the gripper velocity is too high, or the obstacle avoidance scale factor is too small, the grippers can oscillate between servoing to decrease task error, and moving away from obstacles. This effect is most pronounced when the linearizations used inside the controller don't model the local environment well. The second case is when the task is nearly done; in this case if the discretization level of the deformable object or the target points is too coarse, this can lead to rapid changes in the task error gradient, which can cause the controller to oscillate. Last, if stretching correction is directly opposing task progress this will lead to oscillation as the controller switches between the two objectives. The choice of workspace discretization is not critical so long as it is sufficient to capture any relevant details of obstacle geometry.}

\end{document}